\documentclass[sigconf]{acmart}

\AtBeginDocument{%
  }

\usepackage{tabularx}
\usepackage{array}
\usepackage{booktabs}
\usepackage{multirow}
\usepackage[table]{xcolor} % \cellcolor in tables
\usepackage{xstring}       % string test for sign
\usepackage{tablefootnote} % allow \tablefootnote inside tabular
\newcolumntype{Y}{>{\centering\arraybackslash}X}

\definecolor{morandiGreen}{RGB}{206,221,213}
\newcommand{\deltacell}[1]{%
  \IfBeginWith{#1}{-}{\cellcolor{morandiGreen}$#1$}{$#1$}%
}

\usepackage{algorithm}
\usepackage{algpseudocode}

\usepackage{placeins}

\usepackage{graphicx}
\usepackage{subcaption} % subfigures

\usepackage[most]{tcolorbox}
\tcbuselibrary{breakable,skins}
\usepackage{fvextra} % robust verbatim with line-wrapping
\usepackage{capt-of} % \captionof outside floats

\usepackage{microtype}
\usepackage{booktabs}
\usepackage{tabularx}
\usepackage{makecell}
\usepackage{multirow}
\usepackage[table]{xcolor}
\usepackage{array}

\newcolumntype{L}{>{\raggedright\arraybackslash}p{0.20\columnwidth}}

\definecolor{headgray}{HTML}{F6F7F8}
\definecolor{groupgray}{HTML}{EEF0F2}
\definecolor{ourslight}{HTML}{F3F8FF}
\definecolor{oursblue}{HTML}{E6F1FF}
\definecolor{baseyellow}{HTML}{FFF6CC}

\newcommand{\best}[1]{\textbf{#1}}          % best overall
\newcommand{\second}[1]{\underline{#1}}       % second best overall
\newcommand{\basebest}[1]{\cellcolor{baseyellow}{#1}} % best baseline (shaded)
\newcommand{\baseoverall}[1]{\cellcolor{baseyellow}{\textbf{#1}}} % if a baseline is also best overall

\usepackage{booktabs}
\usepackage{siunitx}
\usepackage{multirow}
\usepackage{graphicx}            % \rotatebox
\usepackage[table]{xcolor}
\usepackage{xstring}
\usepackage{footnote}
\makesavenoteenv{table*}         % 让 \tablefootnote 能在 table* 里用

\definecolor{impgreen}{HTML}{1A7A3C}
\definecolor{worsered}{HTML}{B23A48}
\newcommand{\dpct}[1]{%
  \IfBeginWith{#1}{-}%
    {\textcolor{impgreen}{$#1$}}%
    {\textcolor{worsered}{$#1$}}}
\definecolor{lbnlA}{HTML}{E6EEF9}\definecolor{lbnlB}{HTML}{F3F7FC}  % LBNL59 蓝
\definecolor{btsbA}{HTML}{EFE7F8}\definecolor{btsbB}{HTML}{F8F4FC}  % BTS-B 紫
\definecolor{btscA}{HTML}{E6F4EB}\definecolor{btscB}{HTML}{F3FAF5}  % BTS-C 绿    

\newcommand{\dg}{\ensuremath{^{\circ}\mathrm{C}}}   % 摄氏度

\copyrightyear{2026}
\acmYear{2026}
\setcopyright{cc}
\setcctype{by}
\acmConference[SIGSPATIAL '26]{The 34th ACM International Conference on Advances in Geographic Information Systems}{November 03--06, 2026}{Riverside, CA, USA}
\acmBooktitle{The 34th ACM International Conference on Advances in Geographic Information Systems (SIGSPATIAL '26), November 03--06, 2026, Riverside, CA, USA}
\acmDOI{10.1145/3841645.3842954}
\acmISBN{979-8-4007-2950-8/2026/11}

\begin{document}
\raggedbottom % avoid large vertical gaps caused by \flushbottom column stretching

\title{TopoBrick: Agentic Topology Sampling of Exogenous Variables for Zero-Shot Building IoT Forecasting}

\author{Xiachong Lin}
\email{dawn.lin@student.unsw.edu.au}
\orcid{0009-0009-6762-8365}
\affiliation{%
  \institution{University of New South Wales}
  \city{Sydney}
  \state{NSW}
  \country{Australia}
}

\author{Du Yin}
\email{du.yin@unsw.edu.au}
\affiliation{%
  \institution{University of New South Wales}
  \city{Sydney}
  \state{NSW}
  \country{Australia}
}
\author{Arian Prabowo}
\email{arian.prabowo@unsw.edu.au}
\affiliation{%
  \institution{University of New South Wales}
  \city{Sydney}
  \state{NSW}
  \country{Australia}
}

\author{Hao Xue}
\email{haoxue@hkust-gz.edu.cn}
\affiliation{%
 \institution{Hong Kong University of Science and Technology (Guangzhou)}
 \city{Guangzhou}
 \country{China}}

 \author{Wen Hu}
 \email{wen.hu@unsw.edu.au}
\affiliation{%
  \institution{University of New South Wales}
  \city{Sydney}
  \state{NSW}
  \country{Australia}
}

\author{Imran Razzak}
\email{imran.razzak@mbzuai.ac.ae}
\affiliation{%
  \institution{Mohamed bin Zayed University of Artificial Intelligence}
  \city{ Abu Dhabi}
  \country{United Arab Emirates}}

\author{Matthew Amos}
\email{matt.amos@csiro.au}
\affiliation{%
  \institution{CSIRO Energy Centre}
  \city{Newcastle}
  \state{NSW}
  \country{Australia}}

\author{Sam Behrens}
\email{sam.behrens@csiro.au}
\affiliation{%
  \institution{CSIRO Energy Centre}
  \city{Newcastle}
  \state{NSW}
  \country{Australia}}
  
\author{Flora D. Salim}
\email{flora.salim@unsw.edu.au}
\affiliation{%
  \institution{University of New South Wales}
  \city{Sydney}
  \state{NSW}
  \country{Australia}
}

\renewcommand{\shortauthors}{Lin et al.}

\begin{abstract}
Building sensors are embedded in physical topology, spatial hierarchy, and operational context, yet existing forecasters often treat them as isolated time series or rely on fixed covariate sets. 
% This limitation is particularly restrictive for zero-shot forecasting across buildings: although time-series foundation models can transfer without retraining, they lack a principled mechanism to identify which building-specific variables are physically relevant to a given target, while fixed multivariate inputs do not naturally transfer across heterogeneous sensor deployments.
We present \textbf{TopoBrick}, a training-free framework for zero-shot building IoT (Internet-of-Things) forecasting. TopoBrick uses building knowledge graphs to construct a compact structural skeleton and employs an agentic topology sampler to select target-centric exogenous variables. The selected variables are organized by deployment-time availability, separating past-known sensor states from future-known calendar, schedules, and meteorological forecasts. Across three real-world buildings, TopoBrick outperforms strong zero-shot baselines on two of the three and remains competitive with fully trained building-specific models, with the gain concentrated where the routed variables match a real physical coupling. Ablations show that topology-aware sampling is more reliable than random, ontology-only, or fixed-hop selection, especially for physically coupled HVAC and weather-driven sensing variables. The code of this work is available in \url{https://github.com/Dawnlxc/TopoBrick.git}
\end{abstract}

\ccsdesc[500]{Information systems~Spatial-temporal systems}
\ccsdesc[500]{Computing methodologies~Machine learning}
\ccsdesc[300]{Computing methodologies~Knowledge representation and reasoning}
\ccsdesc[300]{Computer systems organization~Sensor networks}
\ccsdesc[100]{Applied computing~Engineering}
\keywords{Building IoT forecasting, building knowledge graph, zero-shot forecasting, topology-aware forecasting, Brick Schema}
%   \includegraphics[width=\textwidth]{samples/sampleteaser}

\maketitle

\section{Introduction}
\label{sec:introduction}

Modern buildings are dense cyber-physical systems instrumented with hundreds to thousands of sensing, control, and operational Points. Forecasting these signals supports demand flexibility, fault detection, predictive control, and grid-interactive operation. Unlike conventional multivariate time-series benchmarks, however, a building does not expose a fixed, semantically aligned feature space. Its Points are embedded in equipment, spaces, air and water loops, and control systems, and the context relevant to a forecast is target-dependent. A zone temperature may depend on its serving equipment, setpoints, and outdoor conditions, whereas an electrical meter may depend on a different subset of the same building. Treating all available channels as interchangeable numerical inputs therefore discards the physical and operational structure that determines which signals are likely to be informative.

Recent time-series foundation models reduce deployment cost by enabling forecasting without building-specific training. However, zero-shot model transfer does not provide zero-shot context construction. In an unseen building, a forecaster still faces hundreds or thousands of heterogeneous Points whose identities, counts, and relationships differ from those in every other deployment. Channel-independent forecasting avoid this problem by independently modeling each target, while channel-dependent models assume that a useful variable set has already been specified. Even when a pretrained forecaster can consume auxiliary variables, it does not have the ability to determine which building-specific variables are physically relevant to a given target or available at inference time. The unresolved question is therefore: \emph{for a target Point in an unseen building, which of the available variables should a zero-shot forecaster condition on?}

Building knowledge graphs (KG) provide a natural cross-building abstraction for this problem.
The semantic schema describes Points through transferable equipment, spatial, and physical-flow relations rather than building-specific channel identities. 
However, a building knowledge graph, which is heterogeneous and hierarchical, is not directly a forecasting graph: most nodes represent equipment, spaces, or systems rather than time series, and relevant Points may be separated by several heterogeneous relations. In addition, KG structures are differentiated by the building blueprint and annotation preference.
Learning a graph encoder would reintroduce the training and schema-alignment requirements that zero-shot deployment is intended to avoid. 
Our central premise is therefore that the knowledge graph should determine what the forecaster sees, rather than act as the forecaster itself: topology should serve as a transferable routing prior for constructing target-centric context.

To this end, we present \textbf{TopoBrick}, a training-free framework that uses the building knowledge graph as a routing layer between an unseen building and a frozen time-series foundation model. 
For each target Point, TopoBrick distills the raw KG into a compact structural skeleton and constructs a target-centric topology context. 
An agentic topology sampler uses this context to propose physically and operationally relevant exogenous variables. The proposals are grounded back to the graph, deterministically materialized into Point time series, and audited by a KG-grounded verifier.
The selected variables are then organized by deployment-time availability: past-known sensor states are restricted to the historical window, whereas future-known calendar, schedule, and meteorological forecasts are available over both the history and prediction horizon. In this formulation, topology is responsible for constraining what information enters the forecaster without parameter updates or future-observation leakage.
Across three real-world buildings, TopoBrick outperforms strong zero-shot baselines in two of the three and remains competitive with fully trained building-specific forecasters, with the largest gains occurring when the building topology exposes physically relevant exogenous variables.

The contributions of this work are summarized as follows:
\begin{itemize}
    \item Formulate zero-shot building sensor forecasting as a topology-aware exogenous-variable selection problem over heterogeneous building knowledge graphs.
    \item Propose TopoBrick, a zero-shot pipeline that constructs compact building skeletons and uses an agentic topology sampler with KG-grounded verification to select target-specific exogenous variables.
    \item Design a deployment-faithful forecasting setup that separates past-known and future-known exogenous variables, and evaluate it on three real-world buildings against zero-shot time series foundation models and fully trained building-specific forecasters.
\end{itemize}

The remainder of this paper is organized as follows. Section~\ref{sec:related_work} reviews related work. Section~\ref{sec:methodology} introduces the proposed methodology. Section~\ref{sec:exp} presents the experimental results. Sections~\ref{sec:discussion} and~\ref{sec:conclusion} provide the discussion and conclusion.

\section{Related Works}
\label{sec:related_work}
\subsection{Forecasting with Exogenous Context}
Time series forecasting has advanced rapidly with deep learning~\cite{xia2025timeemb, yin2024enhancing, yin2025xxltraffic, zhang2023mlpst}, and the resulting architectures differ in ways that matter directly for this work: what they assume about the variables available alongside the target. 
Channel-independent (CI) methods model each series in isolation. DLinear~\cite{zeng2023transformers} showed that a simple linear model can surpass the Transformer architectures that preceded it, including Informer~\cite{zhou2021informer} and its sparse attention over long sequences, redirecting the field from attention mechanisms toward representation. PatchTST~\cite{nie2022time} then demonstrated that patching the input series and treating channels independently substantially improves Transformer performance, FITS~\cite{xu2024fits} and MixLinear~\cite{ma2026mixlinear} reaches comparable accuracy with a fraction of the parameters. All of these models assume the series are independent to avoid cross-variable context modeling.
% Subsequent work questioned the necessity of attention: DLinear~\cite{zeng2023transformers} showed that simple linear models can rival or surpass complex Transformers on standard benchmarks, prompting renewed attention to model design. 
Channel-dependent (CD) models consider the interactions between variables, but assume the exogenous variable sets are fixed and pre-defined across channels.
iTransformer~\cite{liu2024itransformer} inverts the attention mechanism to operate across variables rather than time steps, and TimeMixer~\cite{wang2024timemixer} mixes representations at multiple temporal resolutions. TimeXer~\cite{wang2024timexer} makes the distinction explicit by separating endogenous from exogenous inputs, but the exogenous set is supplied by the user rather than derived. 
A parallel line adapts language models to the modality by reusing a frozen backbone~\cite{zhou2023one, liu2024autotimes}, reprogramming the series into text~\cite{jin2023time, xue2023promptcast}, or extending to multimodal inputs~\cite{jia2024gpt4mts}. All of these require task-specific training.

% More recently, large language models (LLMs) have been adapted for time series. GPT4TS ~\cite{zhou2023one} repurposes a frozen pretrained language model as a general time series backbone, TimeLLM ~\cite{jin2023time} reprograms the input series into the language modality to align with LLM representations, and GPT4MTS ~\cite{jia2024gpt4mts} extends this paradigm to multimodal time series. While effective, these methods generally require task dataset specific training.
Time series foundation models remove the task-specific training requirement.
% A complementary line of work develops time series foundation models that support cross-domain zero-shot forecasting without per-task training. 
Chronos~\cite{ansari2024chronos, ansari2025chronos} tokenizes time series and trains on large-scale corpora to enable zero-shot transfer. TimesFM~\cite{das2023decoder} adopts a decoder-only architecture pretrained on diverse datasets, and Moirai~\cite{woo2024unified, liu2025moirai} unifies arbitrary frequencies and variable counts, each forecasting unseen series zero-shot.
% Despite this progress, these methods are largely developed and evaluated on curated benchmarks where the variable set is fixed and pre-aligned, and they leave two assumptions unaddressed in the building IoT setting. First, channel-independent and implicitly multivariate models offer no principled way to decide \emph{which} of a building's hundreds to thousands of points should serve as exogenous variables for a given target, either ignoring cross-sensor structure or conditioning indiscriminately on all channels, injecting noise and cost. Second, the trained models incur per-building or per-dataset training cost that does not amortize across the long tail of heterogeneous buildings encountered in deployment. Foundation models remove the training cost but still treat exogenous variables as a given flat set, with no notion of the spatial, physical, or control relationships that make an exogenous variable relevant. Our work builds on the zero-shot foundation model paradigm but supplies exactly this missing ingredient: a training-free, building-agnostic mechanism that uses building topology to select physically relevant exogenous variables and to separate past-known from future-known exogenous inputs in real operational settings.
Despite this progress, these methods are largely developed and evaluated on curated benchmarks where the variable set is pre-aligned, leaving two gaps unaddressed in building IoT settings.
First, channel-independent and channel-dependent models offer no principled way to decide which of a building’s hundreds to thousands of points should serve as exogenous variables for a given target, either ignoring cross-sensor structure or conditioning indiscriminately on all channels, thereby introducing noise and computational overhead.
Second, building-specific training incurs substantial computational costs that cannot be amortized across the long tail of heterogeneous buildings encountered in deployment, especially at large IoT scales.
Foundation models remove the training cost but still treat exogenous variables as a given flat set, without accounting for the spatial, physical, or control relationships that determine their relevance to a target.

\subsection{Building KGs and Semantic Analytics}
Building knowledge graphs provide a structured representation of building entities and their relationships. For example, Brick Schema~\cite{balaji2016brick} and Haystack\footnote{\url{https://project-haystack.org/}}, describe equipment, locations, systems, Points, and physical or operational relationships among them. Prior work has used these semantic structures to normalize heterogeneous building metadata, including automatic classification of sensor Points into Brick classes~\cite{prabowo2025brick}. Building semantics are also combined with time series foundation models for building energy analytics~\cite{lin2024bitsa}. For higher-level information access tasks, BuildingQA~\cite{mulayim2025buildingqa} and zero-shot question-answering methods~\cite{mulayim2025towards, lee2026jarvis, ko2026brickqa} use building KGs to ground natural-language queries in building entities and relationships. These studies demonstrate the value of building KGs for semantic normalization, retrieval, and reasoning.

Building time series is separately studied in tasks such as predictions on indoor air quality~\cite{huang2026reliable}, light sensing~\cite{hu2025lightllm}, and electricity load~\cite{xu2022probabilistic} that include privacy-preserving formulations~\cite{zaman2025bloom} and changes in operational patterns under exceptional conditions~\cite{prabowo2023navigating}.
This literature establishes the value of both forecasting and semantic context, but does not address target-specific context construction for a zero-shot forecaster over a heterogeneous building graph. 
% These studies establish both the practical importance of building forecasting and the value of semantic metadata, but they typically treat the knowledge graph as a source of labels or context for a single building, rather than as a structure to be exploited for transferable, cross-building forecasting.
% Critically, none of these lines combine the two capabilities our setting demands. Forecasting methods that use building metadata still rely on building-specific training or handcrafted covariate rules, while KG-centric methods focus on retrieval or question answering rather than driving a forecaster.
% The question of how to reason over a heterogeneous, hierarchical building KG with a large amount of non-TS nodes and varying ontology completeness to select exogenous variables and forecast in a training-free, building-agnostic manner remains open.
Existing forecasting and KG-based approaches therefore stop on opposite sides of the same deployment problem. Forecasting models operate on time series that have already been selected, whereas building-KG methods primarily use topology for semantic annotation, retrieval, or question answering rather than for constructing a forecaster's inputs. TopoBrick connects these two lines of work by treating the building KG as a transferable routing prior rather than as a predictor. Building topology guides the selection of physically and operationally relevant regions, KG evidence grounds the resulting choices in valid Points, and a frozen time series foundation model performs the prediction. The resulting contribution is a training-free, target-specific mechanism for constructing exogenous context across heterogeneous buildings.
\section{Methodology}
\label{sec:methodology}

\begin{figure*}
    \centering
    \includegraphics[width=\linewidth]{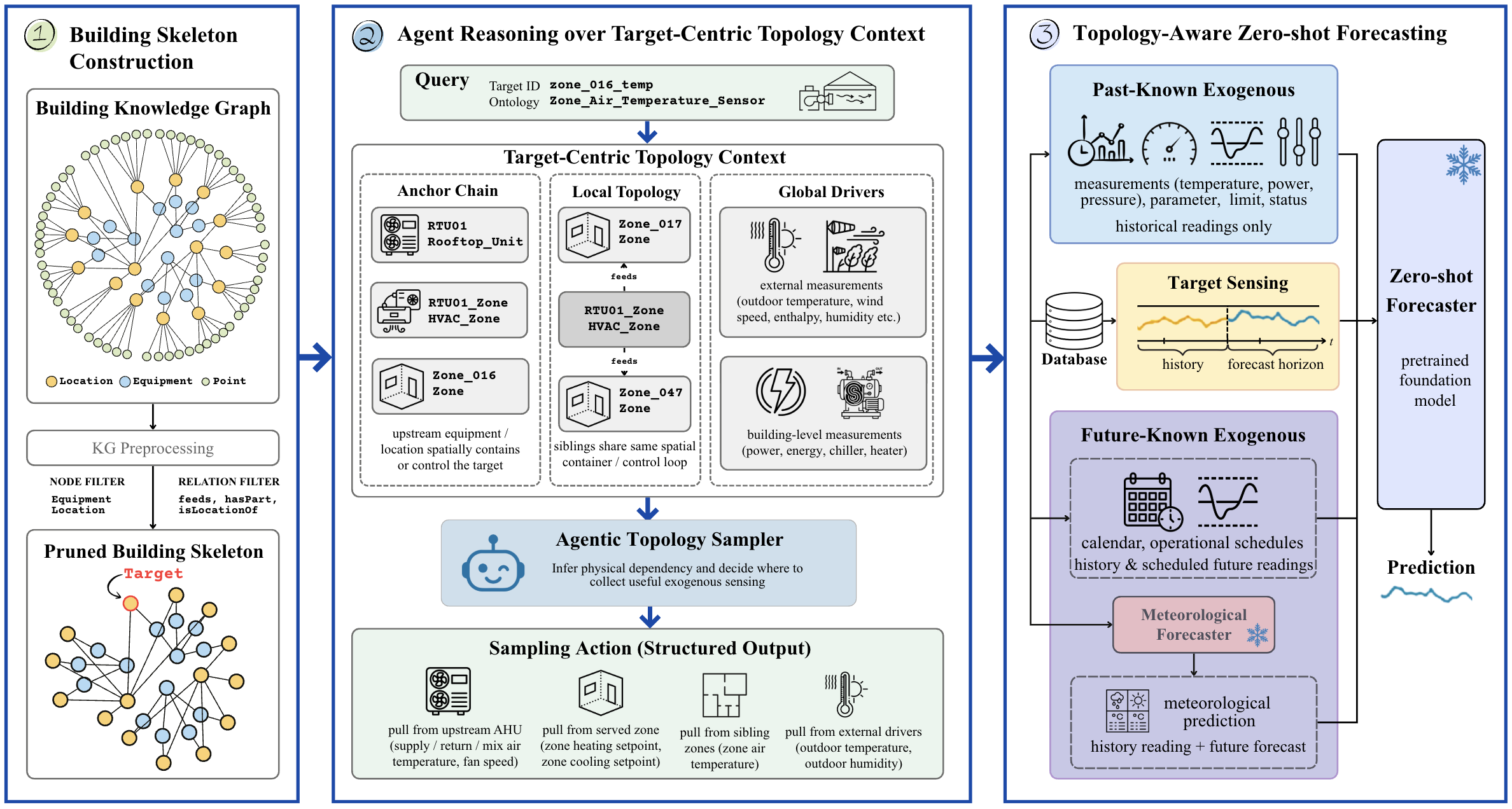}
    \caption{The visualization of the TopoBrick pipeline. (1) Building skeleton construction (2) Agentic reasoning over target-centric topology, selecting physically coupled exogenous variables (3) Topology-aware zero-shot forecaster.}
    \label{fig:pipeline}
\end{figure*}

\paragraph{Overview.} 
% TopoBrick performs zero-shot building sensor forecasting by using the building KG as a routing layer for exogenous variable selection. 
Figure~\ref{fig:pipeline} illustrates the overall pipeline for TopoBrick. Given a target Point node, the framework first constructs a compact building skeleton from the raw KG, then uses an agentic topology sampler to select physically and operationally relevant exogenous variables. The selected variables are verified against KG evidence, materialized into time series inputs, and organized according to their deployment-time availability before being passed to a frozen time series foundation model. The agent is consulted twice per target, once to select and once to audit, and every other step is a deterministic function of the graph. See Alg.~\ref{alg:agentic_subgraph_sampling} for pseudocode.

\paragraph{Notation.} Let $G=(V,E)$ denote a building KG, where $V$ is the KG node set and $E \subseteq V \times \mathcal{R} \times V$ is the set of typed relations. 
Denote $V_{\mathrm{pt}}\subset V$ as the set of Point nodes associated with time series readings, $p\in V_{\mathrm{pt}}$ indicating a target Point, $x^{(p)}$ its observed history, and $y^{(p)}$ the values to be predicted. The exogenous-variable set relevant to $p$ is written as $\mathcal{E}(p)\subseteq V_{\mathrm{pt}}\setminus\{p\}$. In addition, we define $T, S, H$ as the prediction timestep, historical context length, and prediction horizon, respectively.

\subsection{Agentic Topology Subgraph Sampler} 
\label{sec:agentic_sampler} 
The goal of the agentic topology sampler is to use the building KG as a routing layer to identify a set of exogenous variables $\mathcal{E}(p)$ related to the given target Point $p$.
\subsubsection{Building Skeleton Construction}
To support topology-aware reasoning over heterogeneous building KGs, we first pre-process the KG by normalizing relation directions into a canonical top-down form. Inverse relations are converted to their corresponding forward forms according to the Brick ontology.

We define the building skeleton as a reduced topology graph $G_s=(V_s,E_s)$, where $V_s$ contains structural nodes used for topology traversal and $E_s$ contains the structural relations among them:
\begin{align*}
    V_s &= \{\,u\in V \mid \tau(u)\in\mathcal{T}_s\,\}, \\
    E_s &= \{\,(u,r,v)\in E \mid u,v\in V_s,\ r\in\mathcal{R}_s\,\}.
\end{align*}
Here, $\tau(u)$ maps node $u$ to its top-level Brick class, $\mathcal{T}_s$ is the set of structural node types, and $\mathcal{R}_s$ is the set of relations retained for topology traversal, where
\begin{align*}
    \mathcal{T}_s&=\{\texttt{Equipment},\texttt{Location},\texttt{Collection}\}, \\
    \mathcal{R}_s&=\{\texttt{feeds},\texttt{hasPart},\texttt{isLocationOf}\}.
\end{align*}
Nodes carrying time series readings are excluded from $V_s$ even when their ontology types are structural. Instead, Point nodes are maintained as attachments to skeleton nodes. For a Point $p$, we define its set of structural anchors as
\begin{align*}
    A(p)=\{u\in V_s \mid (u,\texttt{hasPoint},p)\in E\}.
\end{align*}
The skeleton therefore separates topology reasoning from Point materialization: the agent first reasons over structural nodes and subsequently expands selected anchors into time-series Points.
\subsubsection{Target-centric Topology Context} 
For each target Point $p$, we construct a target-centric topology context $\mathcal{C}(p)$ from the building skeleton. This context summarizes where the target is located in the building and which structural regions may contain useful exogenous variables. It contains three components:

\textit{Anchor Chain.} The structural path upward from the target Point: the equipment or spatial container it attaches to, and the containers above that one up to the roots the skeleton reaches. This locates the target by its physical position rather than by graph distance, and it is the path along which upstream drivers are found. 

\textit{Local Topology.} The sideways neighborhood at each link of that chain: for every branch above the anchor, the same-class peer cohort that the branch groups. Peers grouped by a common parent typically share a physical subsystem or control loop, so a cohort can be requested as a unit without the agent traversing the full KG. 

\textit{Global Context.} Summarization for global context such as external meteorology monitoring, aggregated demand, or energy usage. This context may not appear in the anchor chain but is globally relevant to the given target.

The topology context is rendered as a compact textual representation for the agent. Instead of exposing raw URI strings or the full KG, the representation uses semantic node types, relation names, Point classes, and aggregate counts, avoiding building-specific naming conventions and allowing the agent to reason over building structure at the appropriate abstraction level. 
% \textit{Target anchor.} The target anchor is the equipment, room, zone, or system node to which the target Point is attached. Points co-located on the same anchor often share a physical subsystem or control loop. 
% \textit{Local topology.} The local topology describes the structural neighborhood around the target anchor: the branches through which the anchor is fed or contained, the same-class peer cohort that each branch groups, and the upstream chain from each branch to the top of the skeleton. This provides physical context without requiring the agent to traverse the full KG. 
% \textit{Global context.} The global context summarizes building-level drivers, such as weather-related Points, utility and building-level meters, and central plant equipment. These variables may not be close to the target under graph distance, but can still be globally relevant. The topology context is rendered as a compact textual representation for the agent. Instead of exposing raw URI strings or the full KG, the representation uses semantic node types, relation names, Point classes, and aggregate counts. This reduces dependence on dataset-specific naming conventions and allows the agent to reason over building structure at the appropriate abstraction level. 
\subsubsection{Exogenous-variable Selection by Agent} Given the target-centric topology context $\mathcal{C}(p)$, the agent selects exogenous variables by reasoning over both target semantics and topology. The reasoning follows three principles:

\textit{Physical Relevance.} The agent first identifies the physical quantity measured by the target Point and infers plausible driver categories. For example, temperature targets may depend on supply or return air temperature, outdoor conditions, setpoints, and equipment states. Pressure and airflow targets may depend on fans, dampers, valves, and loop states. Electrical targets may depend on power, current, voltage, demand, and building-level load conditions. 

\textit{Topological Grounding.} The agent then grounds these driver categories in the skeleton. Rather than retrieving Points solely by ontology class or fixed graph distance, it selects variables from structurally meaningful regions, such as the target anchor, upstream equipment, served zones, sibling components, or building-level hosts. 

\textit{Controlled Expansion.} Finally, the agent determines how broadly each structural region should be expanded. Local variables are retrieved from the target anchor or nearby equipment, whereas global variables are retrieved only when the corresponding physical quantity is building-wide by nature, such as weather or aggregate demand. This avoids uncontrolled expansion from high-degree containers. The agent outputs a structured set of sampling actions $\mathcal{A}(p)=\{(a_i,\omega_i,q_i)\}_{i=1}^{m}$, where $a_i\in V_s$ is a skeleton anchor drawn from $\mathcal{C}(p)$, $q_i$ is the requested Point class, and $\omega_i$ is one of four expansion scopes: the anchor itself, the subtree the anchor contains, the whole building for a class that is building-global by nature, or, when the target has no skeleton anchor, the cluster of Points recovered around it. These actions provide an auditable interface between agent reasoning and deterministic retrieval.

\subsection{Physics Consistency Verifier}
\label{sec:physics_verifier}
\subsubsection{Exogenous-variable Materialization} The actions produced by the sampler are executed by a deterministic materialization module. For each action $(a_i,\omega_i,q_i)\in\mathcal{A}(p)$, the module resolves the anchor $a_i$ against the anchor table constructed together with $\mathcal{C}(p)$, expands the corresponding skeleton region under the scope $\omega_i$, retrieves the attached Point nodes whose Brick class matches $q_i$, removes the target Point itself, and deduplicates repeated selections. This yields the candidate set $\mathcal{L}(p) = \mathrm{Expand}(\mathcal{A}(p))$, where $\mathrm{Expand}$ collects the Points every action resolves to.

Resolution is a table lookup rather than a generation step, which places structural validity outside the agent's reach. An anchor that does not appear in $\mathcal{C}(p)$ has no entry, and a Brick class with no instance under the requested scope returns nothing. The agent selects among regions the knowledge graph exposes, and the expansion of a region is fixed by $G_s$ rather than by the agent.

\begin{table}[t]
\centering
\caption{Rubric used by the physics-consistency verifier.}
\label{tab:verifier_rubric}
\small
\renewcommand{\arraystretch}{1.1}
\begin{tabular}{@{}p{0.24\linewidth}p{0.68\linewidth}@{}}
\toprule
\textbf{Dimension} & \textbf{Criterion} \\
\midrule
Peer cohort &
Checks whether same-ontology spatial peers from the target's serving or nearby equipment are included when available. \\
\addlinespace
Domain drivers &
Checks whether physically meaningful drivers of the target quantity are present. \\
\addlinespace
Noise control &
Flags off-domain variables without a plausible coupling path and overly broad same-class dumps that should be trimmed. \\
\bottomrule
\end{tabular}
\end{table}

\subsubsection{KG-grounded Verification} A second agent audits the composition of $\mathcal{L}(p)$ against the physics of the target. It receives the topology context $\mathcal{C}(p)$ together with the Brick classes present in $\mathcal{L}(p)$ and their counts, and returns a set of classes to remove and a set of additional actions to execute:
\[ (\mathcal{D},\mathcal{A}^{+}) = \mathrm{Verify}\big(\mathcal{C}(p),\ \mathrm{classes}(\mathcal{L}(p))\big). \]
Table~\ref{tab:verifier_rubric} states the rubric: whether the target's same-class spatial peers are represented, whether the physical drivers of the target quantity are present, and whether any class lacks a plausible coupling path to the target or floods the set with a single ontology. Removals are applied before additions, and the additional actions are materialized against the retained set, so a class the audit removes cannot re-enter through its own additions:
\[ \mathcal{E}(p) = \big\{\,\ell\in\mathcal{L}(p) \mid \mathrm{class}(\ell)\notin\mathcal{D}\,\big\} \;\cup\; \mathrm{Expand}(\mathcal{A}^{+}), \]
truncated at $n_{\max}=60$ variables. Auditing classes rather than individual Points keeps the audit at the level at which building physics is stated: that a discharge air temperature is driven by its upstream coil and valve determines which ontologies belong in the set, not which of several interchangeable peer sensors is retained. When $\mathcal{E}(p)$ is instantiated as time series, its Points are intersected with the sensor-level usability audit of Appendix~\ref{appendix:preprocess}, so a routed variable is never a sensor the quality audit has retired. Algorithm~\ref{alg:agentic_subgraph_sampling} in Appendix~\ref{appendix:sampler_algorithm} states the two stages end to end. Section~\ref{sec:topology_aware_forecasting} defines how $\mathcal{E}(p)$ is organized by deployment-time availability.

\subsection{Zero-shot Forecasting with Topology-aware Exogenous Variables} \label{sec:topology_aware_forecasting} 
TopoBrick then formats the selected exogenous variables as inputs to a frozen zero-shot forecaster based on their availability at inference time.
% The key requirement is deployment-time availability: a variable can only be used over the prediction horizon if its future values will be available at inference time. 

\subsubsection{Exogenous-variable Availability} 
\label{sec:exo_avail}
The exogenous variables are divided into \textit{past-known} and \textit{future-known} exogenous variables.
\begin{itemize}
    \item \textit{Past-known exogenous variables:} Including the observations of sensor measurements and equipment states, such as power demand, flow rates, pressures, valve or damper positions, operating modes, and status that are observed up to time $T$, providing historical context but are masked over the prediction horizon. 
    \item \textit{Future-known exogenous variables:} Varibles that are available over both the historical window and the prediction horizon, including
    \begin{itemize}
        \item \textit{Calendar features:} Deterministic functions of timestamps.
        \item \textit{Operational schedules:} HVAC schedules or setpoint programs, which are known in advance from the building management system.
        \item \textit{Meteorological Forecasts:} Observed historically and forecast over the horizon by a separate meteorological forecaster (see Section~\ref{sec:met_forecaster} for details).
    \end{itemize}
\end{itemize}
% \textit{Past-known exogenous variables} are observed only up to the time $T$, including points like sensor measurements, equipment states, power demand, flow rates, pressures, valve positions, damper positions, operating modes, and status variables. They provide historical context but are masked over the prediction horizon. 
% \textbf{Future-known exogenous variables} are available over both the historical window and the prediction horizon, including 
% \begin{itemize}
%     \item \textit{Calendar features:} deterministic functions of timestamps.
%     \item \textit{Operational schedules:} HVAC schedules or setpoint programs, which are known in advance from the building management system.
%     \item \textit{Meteorological Forecasts:} observed historically and forecast over the horizon by a separate weather forecaster.
% \end{itemize}
\subsubsection{Meteorological Forecaster} 
\label{sec:met_forecaster}
Meteorology is an important future-known signal for building IoT forecasting, but ground-truth observations are only available up to the prediction time $T$. Therefore, we introduce a meteorological forecaster that provides meteorological forecasts over the prediction horizon. Denote the historical meteorological readings as
$\mathcal{M}_{T-S+1:T} \in \mathbb{R}^{S\times d_m}$,
where $d_m$ is the number of meteorological variables. The meteorological forecaster maps the given history to future predictions $\widehat{\mathcal{M}}_{T+1:T+H}\in \mathbb{R}^{H\times d_m}$, written as 
\[ \widehat{\mathcal{M}}_{T+1:T+H} = \mathcal{F}_m \left( \mathcal{M}_{T-S+1:T} \right).\] 
% where \[ \widehat{\mathbf{W}}_{T+1:T+H} = [\widehat{\mathbf{w}}_{T+1},\ldots,\widehat{\mathbf{w}}_{T+H}] \in \mathbb{R}^{H\times d_w}. \] 
The produced predictions are therefore concatenated with other future-known variables to form the future-known exogenous variable set when relevant. 
% The meteorological forecaster observes only data available up to time $T$, so no future weather observations are leaked into the target forecasting task. 
\subsubsection{Topology-aware Zero-shot Forecasting} 
Given a target Point $p$ and its history observations $x^{(p)}_{T-S+1:T}\in \mathbb{R}^{S}$, the prediction task is aimed at forecasting the future sequence of $p$, written as $\widehat{y}^{(p)}_{T+1:T+H} \in \mathbb{R}^{H}$.
The selected exogenous variable $\mathcal{E}(p)$ for given $p$ is converted into two aligned matrices following Section~\ref{sec:exo_avail}: 
past-known and future-known exogenous variables, which are written as 
$$X^{past,(p)}_{T-S+1:T} \in \mathbb{R}^{S\times d_p}, \quad X^{future,(p)}_{T-S+1:T+H} \in \mathbb{R}^{(S+H)\times d_f}, $$ respectively. $d_p$ and $d_f$ indicate the numbers of past-known and future-known exogenous variables selected for $p$. 
TopoBrick then performs zero-shot forecasting using a frozen pretrained forecaster $\mathcal{F}_{\theta}$, written as
\[ \widehat{y}^{(p)}_{T+1:T+H} = \mathcal{F}_{\theta} \left( x^{(p)}_{T-S+1:T}, X^{past,(p)}_{T-S+1:T}, X^{future,(p)}_{T-S+1:T+H} \right), \] 
which ensures that the only building-specific computation is on metadata inference when the building is freshly registered.

\section{Experiments}
\label{sec:exp}
\paragraph{Datasets}
We evaluate on comprehensive building operational data spanning three real-world buildings: one campus building \textit{LBNL59} from UC Berkeley, USA~\cite{luo2022three}, and two institutional buildings from Australia, \textit{BTS-B} and \textit{BTS-C}~\cite{prabowo2024building}. Each building is accompanied by a building knowledge graph that describes the semantic connections within the building, together with sensor readings collected from the installed IoT sensors. Training, validation, and testing sets are separated based on the natural calendar (see Table~\ref{tab:dataset_split}). Following the preprocessing procedure in Appendix~\ref{appendix:preprocess}, 103 points in LBNL59, 57 points in BTS-B, and 619 points in BTS-C are selected as evaluated forecasting targets. Data granularity is processed to 15-min, where historical length $S=96$, and prediction horizon $H=\{24, 48, 72, 96\}$, representing 6h, 12h, 18h, 24h ahead multi-step forecasting.

\begin{table}[t]
\centering
\caption{The buildings used for experiments. $G=(V,E)$ is the raw building KG
and $G_s=(V_s,E_s)$ the skeleton derived from it. Each building is split by natural calendar for train/val/test.}
\label{tab:dataset_split}
\small
\setlength{\tabcolsep}{2.5pt}
\begin{tabular*}{\columnwidth}{@{}l@{\extracolsep{\fill}}ccl@{}}
\toprule
 & $|V|/|E|$ & $|V_s|/|E_s|$ & \textbf{Train / Val / Test} \\
\midrule
LBNL59 & 402/499 & 134/225 & May--Jul/Aug/Sep--Oct 2020 \\
BTS-B  & 1{,}115/1{,}368 & 263/470 & Nov--Jan/Feb/Mar--Apr 2021--22 \\
BTS-C  & 11{,}905/10{,}738 & 1{,}445/2{,}537 & Nov--Jan/Feb/Mar--Apr 2022--23 \\
\bottomrule
\end{tabular*}
\end{table}

% \begin{table}[!htb]
%     \centering
%     \begin{tabular}{l|c c c c c }
%     \toprule
%          &  $N$ & $N_{\text{Location}}$ & $N_{\text{Equipment}}$ & $N_{\text{Point}}$\tablefootnote{$N_{\text{Point}}$ denotes the number of unique Brick point entities in the KG (i.e., distinct sensors/meters).} & $N_{\text{Collection}}$\tablefootnote{$N_{\text{Collection}}$ denotes the number of time series collections (e.g., files/streams) associated with points in the dataset.} \\
%          \midrule
%          LBNL59 & 402 & &  \\
%          BTS-A & 1115 \\
%          BTS-B & 11905 \\
%          BTS-C & 9451 \\
%     \bottomrule
%     \end{tabular}
%     \caption{Data summarization}
%     \label{tab:data_summary}
% \end{table}

\paragraph{Baselines}
We compare against three groups of baselines. \textit{Naive baselines} include Persistence that repeats the last observed value, and Seasonal Naive that repeats the value from the corresponding timestamp in the previous seasonal cycle. 
\textit{Full-shot forecasters} include channel-independent methods as DLinear~\cite{zeng2023transformers}, PatchTST~\cite{nie2022time}, 
FITS~\cite{xu2024fits}, and channel-dependent methods
iTransformer~\cite{liu2024itransformer} and TimeXer~\cite{wang2024timexer}. \textit{Zero-shot forecasters} include Chronos-2~\cite{ansari2025chronos}, Moirai~\cite{liu2025moirai}, and TimesFM~\cite{das2023decoder}.

\paragraph{Metrics}
The model is evaluated in mean squared error (MSE) and mean absolute error (MAE). Since building IoT sensors can have different physical units and value ranges, we report normalized metrics for aggregate comparison across sensors. Raw metrics are additionally reported for ontology-specific analysis, where sensors are grouped by Brick class.

\paragraph{Settings}
All supervised baselines are trained with three random seeds, and the average performance among the three are reported. As multivariate forecasters, iTransformer and TimeXer receive the full set of forecastable (See Appendix~\ref{appendix:preprocess}) sensor channels within each building, whereas the remaining univariate baselines operate on individual target series. More implementation details for supervised methods are in Appendix~\ref{appendix:impl_details}. Naive and zero-shot methods are deterministic and are therefore reported as single runs. All methods are evaluated on identical forecasting windows. Calendar features are default applied for all the zero-shot forecasters. Agentic reasoning in TopoBrick is performed using \texttt{gpt-oss-20b}. All experiments are conducted on NVIDIA H100 GPU.

% Per-model learning rates, batch sizes and schedules follow each model's published ETT-minute configuration and are given in the code release. 
% Each is trained under a masked MSE loss on the per-sensor normalized scale the metric is read on, with gradient clipping at $\|\cdot\|_2\!\le\!1$, early stopping on the validation split, and the best checkpoint restored before the test pass. iTransformer and TimeXer are multivariate by construction and receive the building's full channel set rather than the target series alone, and a sensor with no valid window at a given timestamp is excluded from their attention. Every supervised baseline is run with $\mathrm{seed}\in\{2024,2025,2026\}$ and we report the mean. The naive and zero-shot methods are deterministic on a fixed window set and are reported as single values. All methods are scored on the identical evaluation windows.
% All the agentic reasonings are completed by \texttt{gpt-oss-20b}. All experiments are completed on Nvidia H100 GPUs.
\subsection{Main Results}

\begin{table}[!htb]
\centering
\scriptsize
\setlength{\tabcolsep}{3.0pt}
\renewcommand{\arraystretch}{1.03}
\caption{Normalized errors across datasets and prediction horizons. Lower is better. Best overall results are in bold, second-best underlined, and the best baseline shaded. Supervised baselines are averaged over three training seeds.}
\begin{tabularx}{\columnwidth}{@{}L c YY@{\hspace{4pt}}YY@{\hspace{4pt}}YY@{}}
\toprule
\rowcolor{headgray}
\textbf{Method} & \textbf{$H$}
& \multicolumn{2}{c}{\textbf{LBNL59}}
& \multicolumn{2}{c}{\textbf{BTS-B}}
& \multicolumn{2}{c}{\textbf{BTS-C}} \\
\cmidrule(lr){3-4}\cmidrule(lr){5-6}\cmidrule(l){7-8}
\rowcolor{headgray}
& & \textbf{nMAE} & \textbf{nMSE}
& \textbf{nMAE} & \textbf{nMSE}
& \textbf{nMAE} & \textbf{nMSE} \\
\midrule
\rowcolor{groupgray}
\multicolumn{8}{@{}l}{\textbf{Naive Methods}} \\
\multirow{4}{*}{Persistence}
& 24 & 0.653 & 2.016 & 0.435 & 0.746 & 0.417 & 0.860 \\
& 48 & 0.961 & 3.907 & 0.698 & 1.237 & 0.621 & 1.584 \\
& 72 & 1.108 & 4.846 & 0.802 & 1.405 & 0.718 & 1.521 \\
& 96 & 1.075 & 4.320 & 0.752 & 1.230 & 0.707 & 1.499 \\
\cmidrule{1-8}
\multirow{4}{*}{SNaive}
& 24 & 0.926 & 4.103 & 0.449 & 0.715 & 0.550 & 1.103 \\
& 48 & 0.922 & 3.847 & 0.447 & 0.660 & 0.560 & 1.212 \\
& 72 & 0.922 & 3.749 & 0.447 & 0.644 & 0.567 & 0.891 \\
& 96 & 0.914 & 3.540 & 0.441 & 0.569 & 0.569 & 0.902 \\
\midrule
\rowcolor{groupgray}
\multicolumn{8}{@{}l}{\textbf{Full-shot Methods}} \\
\multirow{4}{*}{PatchTST}
& 24 & \baseoverall{0.466} & \basebest{\second{1.117}} & \basebest{\second{0.309}} & \baseoverall{0.413} & \baseoverall{0.305} & 0.578 \\
& 48 & \baseoverall{0.623} & \basebest{\second{1.820}} & \baseoverall{0.383} & \baseoverall{0.478} & \baseoverall{0.399} & \baseoverall{0.813} \\
& 72 & \baseoverall{0.710} & \basebest{\second{2.242}} & \basebest{\second{0.419}} & \baseoverall{0.524} & \baseoverall{0.451} & \baseoverall{0.582} \\
& 96 & \baseoverall{0.751} & \basebest{\second{2.390}} & \basebest{\second{0.429}} & \baseoverall{0.467} & \baseoverall{0.481} & \baseoverall{0.643} \\
\cmidrule{1-8}
\multirow{4}{*}{FITS}
& 24 & 0.588 & 1.704 & 0.396 & 0.487 & 0.386 & 0.634 \\
& 48 & 0.708 & 2.295 & 0.452 & 0.552 & 0.454 & 0.899 \\
& 72 & 0.775 & 2.628 & 0.505 & 0.620 & 0.490 & 0.648 \\
& 96 & 0.798 & 2.664 & 0.550 & 0.618 & 0.508 & 0.692 \\
\cmidrule{1-8}
\multirow{4}{*}{DLinear}
& 24 & 0.493 & 1.250 & 0.320 & \second{0.425} & 0.336 & 0.578 \\
& 48 & 0.639 & 1.970 & 0.396 & \second{0.495} & 0.420 & 0.824 \\
& 72 & 0.720 & 2.366 & 0.423 & 0.531 & 0.464 & 0.604 \\
& 96 & 0.758 & 2.474 & 0.429 & 0.468 & 0.488 & \second{0.655} \\
\cmidrule{1-8}
\multirow{4}{*}{iTransformer}
& 24 & 0.483 & 1.237 & 0.321 & 0.470 & 0.314 & \baseoverall{0.547} \\
& 48 & 0.629 & 1.947 & 0.403 & 0.522 & 0.409 & 0.837 \\
& 72 & 0.725 & 2.397 & 0.431 & 0.555 & 0.468 & 0.612 \\
& 96 & 0.769 & 2.522 & 0.442 & 0.495 & 0.493 & 0.670 \\
\cmidrule{1-8}
\multirow{4}{*}{TimeXer}
& 24 & 0.486 & 1.279 & 0.315 & 0.430 & \second{0.312} & \second{0.555} \\
& 48 & \second{0.626} & 1.937 & 0.404 & 0.504 & \second{0.407} & \second{0.816} \\
& 72 & \second{0.719} & 2.495 & 0.428 & \second{0.528} & \second{0.457} & \second{0.587} \\
& 96 & \second{0.753} & 2.513 & 0.437 & \second{0.467} & \second{0.487} & 0.659 \\
\midrule
\rowcolor{groupgray}
\multicolumn{8}{@{}l}{\textbf{Zero-shot Methods}} \\
\multirow{4}{*}{Moirai2.0}
& 24 & 0.578 & 1.551 & 0.391 & 0.663 & 0.385 & 0.746 \\
& 48 & 0.795 & 2.487 & 0.576 & 0.942 & 0.544 & 1.330 \\
& 72 & 0.901 & 2.967 & 0.633 & 0.964 & 0.622 & 1.157 \\
& 96 & 0.953 & 3.164 & 0.623 & 0.853 & 0.648 & 1.201 \\
\cmidrule{1-8}
\multirow{4}{*}{TimesFM2.0}
& 24 & 0.546 & 1.543 & 0.360 & 0.558 & 0.359 & 0.648 \\
& 48 & 0.749 & 2.595 & 0.494 & 0.755 & 0.508 & 1.170 \\
& 72 & 0.865 & 3.277 & 0.549 & 0.840 & 0.584 & 1.036 \\
& 96 & 0.917 & 3.406 & 0.567 & 0.806 & 0.615 & 1.135 \\
\cmidrule{1-8}
\multirow{4}{*}{Chronos2}
& 24 & 0.519 & 1.351 & 0.316 & 0.510 & 0.326 & 0.669 \\
& 48 & 0.688 & 2.220 & 0.409 & 0.628 & 0.427 & 0.965 \\
& 72 & 0.759 & 2.565 & 0.430 & 0.635 & 0.470 & 0.709 \\
& 96 & 0.793 & 2.619 & 0.433 & 0.568 & 0.495 & 0.752 \\
\cmidrule{1-8}
\rowcolor{oursblue}
& 24 & \second{0.475} & \best{0.976} & \best{0.295} & 0.439 & 0.319 & 0.664 \\
\rowcolor{oursblue}
& 48 & 0.640 & \best{1.691} & \second{0.388} & 0.557 & 0.427 & 0.961 \\
\rowcolor{oursblue}
& 72 & 0.725 & \best{2.064} & \best{0.408} & 0.567 & 0.476 & 0.720 \\
\rowcolor{oursblue}
\multirow{-4}{*}{\makecell[l]{TopoBrick\\}}
& 96 & 0.768 & \best{2.228} & \best{0.413} & 0.505 & 0.502 & 0.779 \\

\bottomrule
\end{tabularx}
\label{tab:main_results}
\end{table}
Table~\ref{tab:main_results} organizes the comparison along two axes: isolating the value of the routed exogenous context by comparing with zero-shot foundation models,  and measuring the performance cost of training-free routing by comparing with full-shot building-specific models. Both axes compare TopoBrick as a whole against methods that route nothing, so neither isolates the sampler from the rest of the pipeline. 
% Section~\ref{sec:exp_sampler} makes that comparison directly, holding the forecaster fixed and varying only how the exogenous set is chosen.

Compared to zero-shot models, Chronos2 is the strongest baseline at every cell on nMAE, and TopoBrick improves on it on two of the three buildings: improving by 3.2\% to 8.5\% on LBNL59 and 4.6\% to 6.6\% on BTS-B. The gains in BTS-C are concentrated at short horizons and slightly hurt the performance in long horizons. Since both TopoBrick and Chronos-2 read the same target history through the same frozen backbone, this isolates the routed context as the source of the difference. The potential causes for this exception are discussed in Section~\ref{sec:discussion}.

Compared to full-shot models, PatchTST as a channel-independent model is the most consistent of the supervised baselines, taking 23 of the 24 cells within the full-shot group, and outperforming all the channel-dependent baselines. 
This result indicates that the breadth of input is not a substitute for selecting a set of physically grounded variables. TopoBrick stays within $7\%$ of PatchTST on nMAE on every building, but which method wins depends on the metric.
On LBNL59, TopoBrick achieves the lowest nMSE at all four horizons ($0.976/1.691/2.064/2.228$), outperforming the strongest baseline by $6.8\%$ to $12.6\%$, while PatchTST holds a $2.0\%$ to $2.7\%$ edge on nMAE. In BTS-B, TopoBrick leads on nMAE at three of four horizons by $2.6\%$ to $4.5\%$, trails PatchTST by $1.3\%$ at $H{=}48$, and sits $6.3\%$ to $16.5\%$ behind on nMSE. On BTS-C, the trained models keep a clear margin on both, $4.4\%$ to $7.0\%$ on nMAE and $18.2\%$ to $23.7\%$ on nMSE.

Considering nMSE squares the residuals, where nMSE and nMAE can respond differently to a minority of large deviations, we therefore examine where each building's nMSE is concentrated, and find that the two buildings differ in kind. On LBNL59 the concentration is temporal. The worst $5\%$ of forecast windows account for $77.0\%$ of the total nMSE at $H{=}24$ and $71.1\%$ at $H{=}96$, and half of these windows fall within 14 of the 61 test days. The affected targets are not a small unreliable subset: 96 of the 103 targets contribute to the tail, and the five largest contributors account for under a quarter of it. The tail therefore corresponds to a building-wide operating condition over a limited period, which the target history alone predicts poorly and on which routed exogenous context is most informative. 
On BTS-B the concentration is by target. Three sensors account for $92\%$ of the nMSE gap to PatchTST at $H{=}24$, where excluding the two largest reduces the remaining gap to $1.4\%$, which is within the reasonable variation. On the two targets where PatchTST is weakest, TopoBrick is the more accurate of the two. The nMSE gap on this building is therefore attributable to a small number of target classes rather than to a systematically heavier error tail. Section~\ref{sec:exp_ontology} identifies these classes and Section~\ref{sec:discussion} examines why topology provides them little useful context.

Across three real-world buildings, TopoBrick is competitive and lands within single-digit percentages of models fully trained on each building, demonstrating that topology-aware exogenous selection with a deployment-faithful organization offers a practical alternative to supervised forecasting for building IoT.

\subsection{Performance Analysis by Ontology}
\label{sec:exp_ontology}

Table~\ref{tab:per-ontology-analysis} analyzes performance by ontology in each building. We use the Chronos-2 model where calendar features are provided as default future-known exogenous variables as the compared baseline for computing the relative raw MAE change $\Delta$ for each ontology. In addition, the number of sensors and raw MAE by ontology are also tabulated for reference. \textcolor{impgreen}{Green}/\textcolor{worsered}{red} indicates improvement/degradation compared to the baseline.

%%%% DU 
\begin{table*}[p]
\centering
\small
\caption{Per-ontology performance by building: raw MAE (rMAE) and relative change $\Delta\%$ (baseline$\rightarrow$model) at 24/48/72/96 steps. }
\renewcommand{\arraystretch}{0.91}
\begin{tabular}{@{}c l r c *{4}{S[table-format=3.3] c}@{}}
\toprule
 & \multirow{2}{*}{\textbf{Ontology}} & \multicolumn{2}{c}{\textbf{Horizon}}
 & \multicolumn{2}{c}{\textbf{24}} & \multicolumn{2}{c}{\textbf{48}}
 & \multicolumn{2}{c}{\textbf{72}} & \multicolumn{2}{c}{\textbf{96}} \\
\cmidrule(lr){5-6}\cmidrule(lr){7-8}\cmidrule(lr){9-10}\cmidrule(lr){11-12}
 & & $\mathbf{N}$ & \textbf{Unit} & {rMAE} & {$\Delta\%$} & {rMAE} & {$\Delta\%$} & {rMAE} & {$\Delta\%$} & {rMAE} & {$\Delta\%$} \\
\midrule
\rowcolor{lbnlA} & \texttt{Zone\_Air\_Temperature\_Sensor} & 67  & \dg\footnote{The unit of \texttt{Zone\_Air\_Temperature\_Sensor} in LBNL59 are mixed with $^{\circ}\mathrm{F}$ and $^{\circ}\mathrm{C}$. We convert all $^{\circ}\mathrm{F}$ readings to $^{\circ}\mathrm{C}$ for convenient.} & 0.184 & \dpct{-6.3} & 0.259 & \dpct{-3.4} & 0.296 & \dpct{-0.2} & 0.316 & \dpct{+0.8} \\
\rowcolor{lbnlB} & \texttt{CO2\_Sensor}            & 11 & ppm & 6.237 & \dpct{-5.7}  & 7.692 & \dpct{-5.3}  & 8.267 & \dpct{-4.5}  & 8.630 & \dpct{-3.6} \\
\rowcolor{lbnlA} & \texttt{Building\_Electrical\_Meter\_Sensor}      & 5  & kW  & 1.031 & \dpct{-2.6}  & 1.240 & \dpct{-0.6}  & 1.399 & \dpct{+0.1}  & 1.562 & \dpct{-0.3} \\
\rowcolor{lbnlB} & \texttt{Mixed\_Air\_Temperature\_Sensor}          & 4  & \dg & 0.537 & \dpct{-10.1} & 0.798 & \dpct{-8.8}  & 0.959 & \dpct{-6.1}  & 1.017 & \dpct{-5.9} \\
\rowcolor{lbnlA} & \texttt{Return\_Air\_Temperature\_Sensor}         & 4  & \dg & 0.227 & \dpct{-10.8} & 0.317 & \dpct{-9.9}  & 0.359 & \dpct{-9.6}  & 0.370 & \dpct{-8.4} \\
\rowcolor{lbnlB} & \texttt{Supply\_Air\_Temperature\_Sensor}         & 4  & \dg & 0.243 & \dpct{-22.8} & 0.355 & \dpct{-21.9} & 0.429 & \dpct{-20.2} & 0.467 & \dpct{-17.7} \\
\rowcolor{lbnlA} & \texttt{Chilled\_Water\_Temperature\_Sensor}      & 2  & \dg & 0.701 & \dpct{-3.2}  & 0.870 & \dpct{-1.4}  & 0.944 & \dpct{-0.5}  & 1.008 & \dpct{+0.1} \\
\rowcolor{lbnlB} & \texttt{Outdoor\_Air\_Temperature\_Sensor}        & 2  & \dg & 1.569 & \dpct{-2.5}  & 2.184 & \dpct{-3.0}  & 2.541 & \dpct{-2.3}  & 2.753 & \dpct{-2.6} \\
\rowcolor{lbnlA} & \texttt{Outdoor\_Air\_Humidity\_Sensor}           & 1  & \%  & 6.777 & \dpct{+0.4}  & 10.490 & \dpct{+1.2} & 12.912 & \dpct{+3.7} & 14.038 & \dpct{+3.9} \\
\rowcolor{lbnlB}\multirow{-10}{*}{\rotatebox{90}{\textbf{LBNL59}}} & \texttt{Hot\_Water\_Temperature\_Sensor}        & 2  & \dg & 0.627 & \dpct{+1.0}  & 0.771 & \dpct{+1.2}  & 0.910 & \dpct{+0.7}  & 0.944 & \dpct{+0.4} \\
\midrule
\rowcolor{btsbA} & \texttt{Air\_Temperature\_Sensor} & 23& \dg & 0.549 & \dpct{-1.2}  & 0.794  & \dpct{+1.9}  & 0.841  & \dpct{+1.7}  & 0.869  & \dpct{+2.7} \\
\rowcolor{btsbB} & \texttt{Discharge\_Air\_Temperature\_Sensor}              & 10 & \dg & 0.922 & \dpct{+0.8}  & 1.234  & \dpct{+2.9}  & 1.345  & \dpct{+3.1}  & 1.380  & \dpct{+3.1} \\
\rowcolor{btsbA} & \texttt{Filter\_Differential\_Pressure\_Sensor}           & 5  & --  & 8.496 & \dpct{+2.9}  & 18.804 & \dpct{+2.0}  & 22.603 & \dpct{-0.9}  & 19.887 & \dpct{-2.1} \\
\rowcolor{btsbB} & \texttt{Average\_Zone\_Air\_Temperature\_Sensor}           & 2  & \dg & 0.442 & \dpct{-3.4}  & 0.650  & \dpct{-3.3}  & 0.730  & \dpct{-0.6}  & 0.724  & \dpct{+0.6} \\
\rowcolor{btsbA} & \texttt{Differential\_Pressure\_Sensor}                  & 2  & --  & 2.272 & \dpct{-79.2} & 2.821  & \dpct{-78.3} & 3.283  & \dpct{-71.8} & 3.513  & \dpct{-67.4} \\
\rowcolor{btsbB} & \texttt{Discharge\_Water\_Temperature\_Sensor}            & 2  & \dg & 4.651 & \dpct{-33.2} & 5.082  & \dpct{-36.4} & 4.952  & \dpct{-36.8} & 4.927  & \dpct{-35.0} \\
\rowcolor{btsbA} & \texttt{Return\_Water\_Temperature\_Sensor}               & 2  & \dg & 3.599 & \dpct{-35.9} & 4.043  & \dpct{-37.2} & 4.280  & \dpct{-36.0} & 4.372  & \dpct{-32.7} \\
\rowcolor{btsbB} & \texttt{Warmest\_Zone\_Air\_Temperature\_Sensor}           & 2  & \dg & 0.522 & \dpct{-3.1}  & 0.746  & \dpct{-1.8}  & 0.802  & \dpct{+4.7}  & 0.894  & \dpct{+7.8} \\
\rowcolor{btsbA} & \texttt{Chilled\_Water\_Differential\_Temperature\_Sensor} & 1  & \dg & 0.609 & \dpct{-1.5}  & 0.785  & \dpct{-0.6}  & 0.879  & \dpct{+0.3}  & 0.901  & \dpct{-1.1} \\
\rowcolor{btsbB} & \texttt{Outside\_Air\_Temperature\_Sensor} & 6 & \dg & 1.426 & \dpct{+1.7} & 1.851 & \dpct{+0.7} & 1.987 & \dpct{+0.9} & 2.073 & \dpct{+1.0} \\
\rowcolor{btsbA} & \texttt{Outside\_Air\_Humidity\_Sensor} & 1 & \% & 6.896 & \dpct{+8.2} & 8.739 & \dpct{+4.7} & 9.373 & \dpct{+2.9} & 9.868 & \dpct{+2.9} \\
\rowcolor{btsbB}\multirow{-12}{*}{\rotatebox{90}{\textbf{BTS-B}}} & \texttt{Water\_Temperature\_Sensor}                & 1  & \dg & 0.627 & \dpct{-1.6}  & 0.675  & \dpct{+5.7}  & 0.807  & \dpct{-1.3}  & 0.713  & \dpct{+2.9} \\
\midrule
\rowcolor{btscA} & \texttt{Electrical\_Power\_Sensor} & 202 & kW\,/\,W & 2.909   & \dpct{+2.8}  & 3.627   & \dpct{+6.4}  & 3.926   & \dpct{+5.8}  & 3.908   & \dpct{+5.1} \\
\rowcolor{btscB} & \texttt{Zone\_Air\_Temperature\_Sensor}              & 67  & \dg & 0.355   & \dpct{-9.8}  & 0.534   & \dpct{-8.0}  & 0.632   & \dpct{-4.2}  & 0.695   & \dpct{-1.8} \\
\rowcolor{btscA} & \texttt{Air\_Temperature\_Sensor}                   & 55  & \dg & 0.482   & \dpct{-7.8}  & 0.663   & \dpct{-4.5}  & 0.726   & \dpct{-4.6}  & 0.764   & \dpct{-4.9} \\
\rowcolor{btscB} & \texttt{Relative\_Humidity\_Sensor}                 & 44  & \%  & 2.933   & \dpct{-8.8}  & 4.302   & \dpct{-9.1}  & 5.119   & \dpct{-6.9}  & 5.608   & \dpct{-5.5} \\
\rowcolor{btscA} & \texttt{Air\_Flow\_Sensor}                          & 26  & --  & 41.952  & \dpct{+0.8}  & 51.548  & \dpct{+2.7}  & 59.933  & \dpct{+2.5}  & 62.457  & \dpct{+2.2} \\
\rowcolor{btscB} & \texttt{Supply\_Air\_Temperature\_Sensor}            & 24  & \dg & 0.767   & \dpct{-6.2}  & 1.082   & \dpct{-0.5}  & 1.241   & \dpct{+2.2}  & 1.331   & \dpct{+3.2} \\
\rowcolor{btscA} & \texttt{Average\_Zone\_Air\_Temperature\_Sensor}      & 23  & \dg & 0.354   & \dpct{-8.0}  & 0.541   & \dpct{-3.4}  & 0.633   & \dpct{+0.0}  & 0.688   & \dpct{+1.2} \\
\rowcolor{btscB} & \texttt{Supply\_Air\_Static\_Pressure\_Sensor}        & 16  & --  & 2.272   & \dpct{+4.3}  & 2.284   & \dpct{+9.2}  & 2.545   & \dpct{+10.6} & 2.728   & \dpct{+11.0} \\
\rowcolor{btscA} & \texttt{Outside\_Air\_Temperature\_Sensor}           & 14  & \dg & 0.655   & \dpct{-40.2} & 0.827   & \dpct{-41.7} & 0.883   & \dpct{-42.4} & 0.915   & \dpct{-42.5} \\
\rowcolor{btscB} & \texttt{Air\_Enthalpy\_Sensor}                      & 12  & --  & 1.187   & \dpct{-33.4} & 1.661   & \dpct{-36.1} & 1.968   & \dpct{-34.7} & 2.167   & \dpct{-33.7} \\
\rowcolor{btscA} & \texttt{Return\_Air\_Humidity\_Sensor}               & 10  & \%  & 2.332   & \dpct{-13.2} & 3.436   & \dpct{-11.2} & 4.024   & \dpct{-8.7}  & 4.399   & \dpct{-7.9} \\
\rowcolor{btscB} & \texttt{Return\_Air\_Temperature\_Sensor}            & 10  & \dg & 0.454   & \dpct{-25.9} & 0.608   & \dpct{-25.0} & 0.690   & \dpct{-22.6} & 0.738   & \dpct{-22.2} \\
\rowcolor{btscA} & \texttt{Temperature\_Sensor}                       & 10  & \dg & 0.811   & \dpct{-5.4}  & 0.960   & \dpct{-5.4}  & 0.855   & \dpct{-3.6}  & 0.882   & \dpct{-0.3} \\
\rowcolor{btscB} & \texttt{Outside\_Air\_Enthalpy\_Sensor}              & 9   & --  & 1.529   & \dpct{-35.7} & 2.133   & \dpct{-35.7} & 2.403   & \dpct{-35.3} & 2.581   & \dpct{-34.9} \\
\rowcolor{btscA} & \texttt{CO2\_Sensor}               & 9   & ppm & 16.511  & \dpct{-8.6}  & 23.081  & \dpct{-4.9}  & 25.593  & \dpct{-3.2}  & 27.718  & \dpct{-1.2} \\
\rowcolor{btscB} & \texttt{Outside\_Air\_Humidity\_Sensor}              & 8   & \%  & 4.499   & \dpct{-17.3} & 6.265   & \dpct{-15.3} & 6.996   & \dpct{-13.2} & 7.456   & \dpct{-11.9} \\
\rowcolor{btscA} & \texttt{Demand\_Sensor}                            & 8   & kW  & 100.086 & \dpct{+10.8} & 104.932 & \dpct{+13.0} & 18.339  & \dpct{+20.6} & 19.013  & \dpct{+18.6} \\
\rowcolor{btscB} & \texttt{Return\_Air\_Enthalpy\_Sensor}               & 7   & --  & 1.329   & \dpct{-19.6} & 1.908   & \dpct{-21.4} & 2.257   & \dpct{-19.2} & 2.455   & \dpct{-19.0} \\
\rowcolor{btscA} & \texttt{Exhaust\_Air\_Temperature\_Sensor}           & 5   & \dg & 0.519   & \dpct{-48.9} & 0.659   & \dpct{-50.1} & 0.712   & \dpct{-50.7} & 0.742   & \dpct{-50.8} \\
\rowcolor{btscB} & \texttt{Exhaust\_Air\_Humidity\_Sensor}              & 5   & \%  & 5.063   & \dpct{-12.8} & 7.129   & \dpct{-11.3} & 8.019   & \dpct{-8.4}  & 8.493   & \dpct{-7.5} \\
\rowcolor{btscA} & \texttt{Zone\_Air\_Humidity\_Sensor}                 & 5   & \%  & 1.662   & \dpct{-4.6}  & 2.508   & \dpct{-4.9}  & 3.020   & \dpct{-3.7}  & 3.321   & \dpct{-3.2} \\
\rowcolor{btscB} & \texttt{Chilled\_Water\_Flow\_Sensor}                & 5   & --  & 0.285   & \dpct{-4.5}  & 0.442   & \dpct{-9.0}  & 0.657   & \dpct{-8.4}  & 0.733   & \dpct{-8.8} \\
\rowcolor{btscA} & \texttt{Discharge\_Air\_Flow\_Sensor}                & 5   & --  & 52.701  & \dpct{+4.5}  & 89.937  & \dpct{+12.7} & 111.443 & \dpct{+15.6} & 123.168 & \dpct{+15.4} \\
\rowcolor{btscB} & \texttt{Supply\_Air\_Humidity\_Sensor}               & 4   & \%  & 4.459   & \dpct{-7.0}  & 6.500   & \dpct{-4.2}  & 7.506   & \dpct{-1.2}  & 8.056   & \dpct{+0.3} \\
\rowcolor{btscA} & \texttt{Enthalpy\_Sensor}                          & 4   & --  & 1.129   & \dpct{-16.5} & 1.749   & \dpct{-17.7} & 2.183   & \dpct{-16.0} & 2.455   & \dpct{-14.6} \\
\rowcolor{btscB} & \texttt{Chilled\_Water\_Supply\_Temperature\_Sensor}  & 4   & \dg & 1.109   & \dpct{-0.3}  & 1.396   & \dpct{-3.9}  & 1.523   & \dpct{-6.8}  & 1.637   & \dpct{-7.6} \\
\rowcolor{btscA} & \texttt{Chilled\_Water\_Return\_Temperature\_Sensor}  & 4   & \dg & 1.260   & \dpct{-4.2}  & 1.761   & \dpct{-0.7}  & 2.051   & \dpct{+1.5}  & 2.245   & \dpct{+4.0} \\
\rowcolor{btscB} & \texttt{Outside\_Air\_Dewpoint\_Sensor}  & 3   & \dg & 0.919   & \dpct{-23.0} & 1.213   & \dpct{-28.5} & 1.348   & \dpct{-29.4} & 1.428   & \dpct{-29.3} \\
\rowcolor{btscA} & \texttt{Dewpoint\_Sensor}      & 3   & \dg & 1.102   & \dpct{-11.1} & 1.523   & \dpct{-8.1}  & 1.728   & \dpct{-5.6}  & 1.820   & \dpct{-4.7} \\
\rowcolor{btscB} & \texttt{Filter\_Differential\_Pressure\_Sensor}      & 3   & --  & 3.444   & \dpct{-15.1} & 4.444   & \dpct{-27.5} & 4.410   & \dpct{-33.2} & 5.278   & \dpct{-30.7} \\
\rowcolor{btscA} & \texttt{Humidity\_Sensor} & 2   & \%  & 4.492   & \dpct{-0.3}  & 7.117   & \dpct{+11.5} & 7.831   & \dpct{+12.5} & 8.259   & \dpct{+12.5} \\
\rowcolor{btscB} & \texttt{Mixed\_Air\_Temperature\_Sensor} & 2   & \dg & 0.596   & \dpct{-10.9} & 0.786   & \dpct{-5.0}  & 0.854   & \dpct{-4.4}  & 0.929   & \dpct{-3.2} \\
\rowcolor{btscA} & \texttt{Hot\_Water\_Return\_Temperature\_Sensor}      & 2   & \dg & 1.969   & \dpct{-0.8}  & 2.497   & \dpct{+2.7}  & 2.390   & \dpct{+2.4}  & 2.233   & \dpct{+3.1} \\
\rowcolor{btscB} & \texttt{Water\_Temperature\_Sensor} & 2   & \dg & 1.300   & \dpct{-0.5}  & 2.263   & \dpct{+1.5}  & 3.165   & \dpct{+2.1}  & 3.923   & \dpct{+0.7} \\
\rowcolor{btscA} & \texttt{Hot\_Water\_Supply\_Temperature\_Sensor} & 2   & \dg & 0.867   & \dpct{+8.9}  & 1.047   & \dpct{+11.2} & 1.123   & \dpct{+11.5} & 1.135   & \dpct{+9.8} \\
\rowcolor{btscB} & \texttt{Chilled\_Water\_Differential\_Pressure\_Sensor} & 2 & -- & 0.667   & \dpct{+10.2} & 0.666   & \dpct{+20.2} & 0.186   & \dpct{+11.6} & 0.149   & \dpct{+6.0} \\
\rowcolor{btscA} & \texttt{Air\_Differential\_Pressure\_Sensor} & 1   & --  & 1.374   & \dpct{-3.6}  & 1.473   & \dpct{-9.6}  & 1.111   & \dpct{-3.3}  & 1.195   & \dpct{-2.4} \\
\rowcolor{btscB} & \texttt{Cooling\_Demand\_Sensor} & 1   & kW  & 0.288   & \dpct{-1.3}  & 0.317   & \dpct{+0.2}  & 0.334   & \dpct{-0.9}  & 0.343   & \dpct{-0.1} \\
\rowcolor{btscA}\multirow{-39}{*}{\rotatebox{90}{\textbf{BTS-C}}} & \texttt{Power\_Sensor} & 1  & kW  & 1.984   & \dpct{-3.5}  & 2.022   & \dpct{-4.0}  & 0.121   & \dpct{-10.0} & 0.123   & \dpct{-8.3} \\
\bottomrule
\end{tabular}

\label{tab:per-ontology-analysis}
\end{table*}

On LBNL59, the improvements are concentrated on air-side temperature variables. In particular, \textit{Supply Air Temperature Sensor} achieves consistent reductions across all horizons, from $-22.8\%$ at 24 steps to $-17.7\%$ at 96 steps. Similar but smaller improvements are observed for \textit{Return Air Temperature Sensor}, \textit{Mixed Air Temperature Sensor}, and \textit{CO2 Sensor}. These results suggest that topology-aware exogenous variables are especially useful when the target signal is governed by HVAC air-loop interactions. In contrast, the improvement on \textit{Zone Air Temperature Sensor} decreases with the prediction horizon and becomes slightly positive at 96 steps, indicating that zone-level temperature may already be well captured by temporal persistence and that additional context is less beneficial when unobserved occupancy or control actions dominate longer-term variations.
BTS-B exhibits a more selective pattern. The strongest improvements are observed for \textit{Differential Pressure Sensor}, \textit{Discharge Water Temperature Sensor}, and \textit{Return Water Temperature Sensor}, with large error reductions across all horizons. This indicates that the sampled context is particularly effective for water-loop and pressure-related variables in this building. However, several air-temperature and outdoor-air variables show limited or negative gains, suggesting that semantic similarity alone is insufficient and that the utility of exogenous variables depends on the underlying building topology and operating regime.
BTS-C provides the most diverse ontology-level evidence. Large and stable improvements are observed for weather-coupled and thermodynamic variables, including \textit{Outside Air Temperature Sensor}, \textit{Outside Air Enthalpy Sensor}, \textit{Air Enthalpy Sensor}, \textit{Exhaust Air Temperature Sensor}, \textit{Return Air Temperature Sensor}, and \textit{Outside Air Dewpoint Sensor}. These classes are directly affected by outdoor conditions, air mixing, and heat exchange processes, making them well suited to topology-aware and future-context-aware modeling. By contrast, \textit{Electrical Power Sensor}, \textit{Demand Sensor}, \textit{Supply Air Static Pressure Sensor}, and \textit{Discharge Air Flow Sensor} show increased error. These signals are often driven by stochastic equipment usage, occupancy, and closed-loop control actions, which may not be fully captured by the available exogenous variables.

The split is sharp enough to read as a rule. Where a target's drivers are themselves instrumented and structurally reachable, routing finds them and the error falls. Where the dominant driver is occupancy, discretionary equipment use, or a supervisory control loop, no Point in the graph stands for it, and the sampler can only return variables that are structurally adjacent without being causally related.

\subsection{Comparison on Sampling Strategy}
\label{sec:exp_sampler}
We compare the proposed agentic sampler against several alternative sampling strategies to validate the information gains from the topology-aware sampler.
% \textit{Random Sampler:}
% This strategy randomly selects building Points as exogenous variables without considering either semantic alignment or topological dependency.
\begin{itemize}
    \item \textit{Random Sampler:} Randomly selects building Points as exogenous variables without considering either semantic alignment or topological dependency.
    \item \textit{Same Ontology Sampler:} Selects points that share the same Brick ontology class with the target sensor as exogenous variables. 
    \item \textit{$k$-hop Sampler:} Selects Points within a fixed $k$-hop neighborhood of the target node as exogenous variables. 
\end{itemize}
% \textit{Same Ontology Sampler:}
% This strategy selects points that share the same Brick ontology class as the target sensor. 
% Although ontology-level matching can quickly retrieve semantically aligned signals, it is blind to topology. In large commercial buildings, sensors with the same ontology class may be distributed across unrelated equipment or locations, which has the possibility of introducing redundant or weakly informative variables.
% \textit{$k$-hop Sampler:}
% Selecting points within a fixed $k$-hop neighborhood of the target node as exogenous variables. 
% It utilizes graph connectivity to capture local structural context. However, all neighboring nodes are treated uniformly, where irrelevant signals could be involved.
To isolate the effect of the sampling strategy, we evaluate all samplers under the same forecaster and compare their error changes against a baseline without considering KG. 
For each sampler, we follow the rule in Section~\ref{sec:exo_avail} to separate the exogenous by availability.
Oracle meteorological forecasts are applied as the future-known exogenous to remove uncertainty from external forecasting errors and ensure that the comparison focuses on the quality of exogenous variable selection.
Similar to Section~\ref{sec:exp_ontology}, calendar features are applied default. 
We report the percentage changes in normalized MAE and MSE, where negative values indicate improvement.

\begin{figure}
    \centering
    \includegraphics[width=\linewidth]{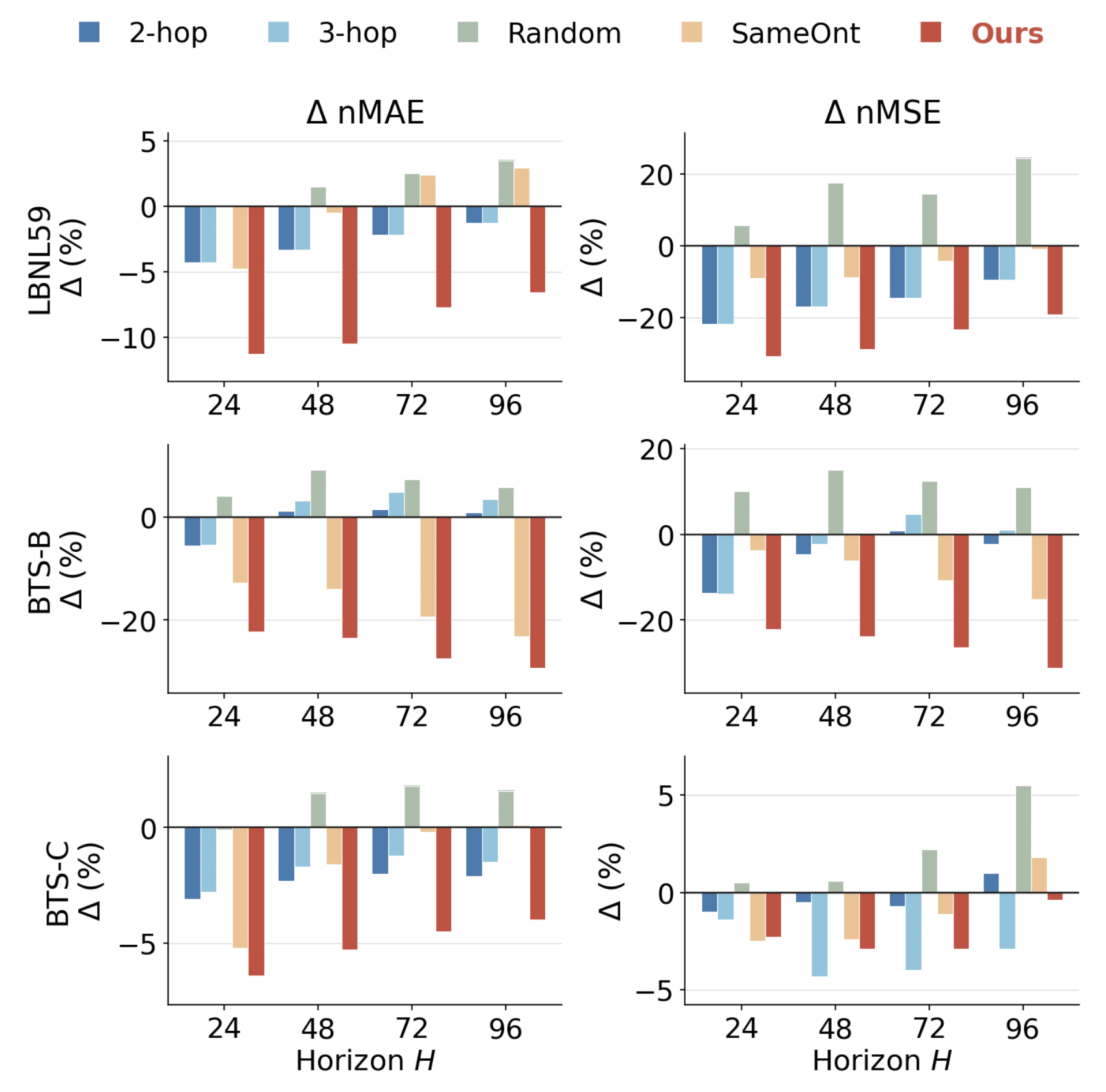}
    \caption{Visualization of the sampler comparison. The x-axis indicates the prediction horizon, and y-axis indicates the $\Delta$ against the baseline. Negative $\Delta$ means performance improvement.}
    \label{fig:sampler_ablation}
\end{figure}

Figure~\ref{fig:sampler_ablation} visualizes the experiment results, which show consistent achievement for the strongest improvements by the proposed agentic sampler. On LBNL59, it gives the largest reduction on both metrics at every horizon, reducing normalized MAE by 2.4 to 5.1 times as much as the strongest alternative and normalized MSE by 1.4 to 2.0 times. 2- and 3-hop sampling score identically here, indicating that merely using local graph neighborhoods is insufficient. Selecting structurally meaningful variables is more important than expanding the neighborhood size.
The advantage of TopoBrick agentic sampler is most pronounced on BTS-B. It leads the strongest alternative at every horizon, reducing normalized MAE by 1.3 to 1.7 times as much and normalized MSE by 1.6 to 3.9 times as much. In contrast, random sampling consistently increases error, confirming that injecting arbitrary building signals can introduce noise rather than useful context. The same-ontology sampler performs competitively on BTS-B, particularly for normalized MAE, but its improvement remains weaker than TopoBrick. This suggests that semantic similarity is useful when same-class sensors share common operational patterns, but it still misses cross-ontology physical dependencies such as the physical coupling in spatial containers or control loops.
On BTS-C, all samplers show smaller gains, suggesting that the benefit of exogenous variables is more dataset-dependent. Nevertheless, TopoBrick remains the best or among the best strategies across most horizons, indicating that topology-aware sampling provides robust benefits.

The results support the central hypothesis that building sensors should not be treated as isolated time series. Same-ontology retrieval captures semantic similarity but may select physically unrelated sensors, while fixed $k$-hop sampling captures local connectivity but lacks relevance filtering. TopoBrick provides a better trade-off by selecting exogenous variables that are both structurally connected and operationally meaningful, leading to more reliable forecasting improvements. 
% More details about the results are in ~\ref{appendix:sampler}

\subsection{Ablations Study}
\label{sec:exp_ablation_exo_group}

\subsubsection{Ablation on Future Known Exogenous Variables}
\label{sec:exp_ablation_exo}
\begin{figure}[h!]
    \centering
    \includegraphics[width=\linewidth]{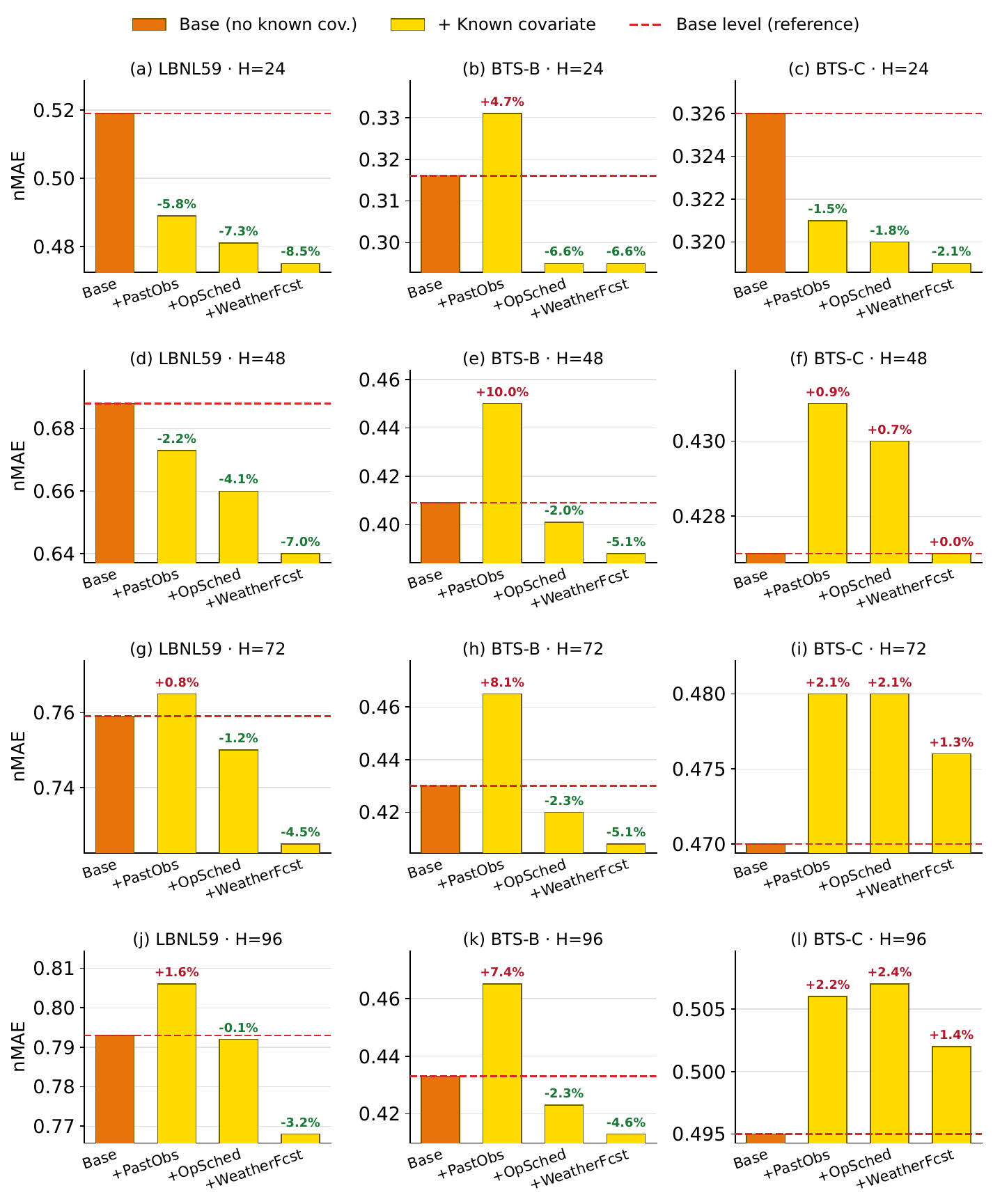}
    \caption{The average normalized MAE (nMAE) of availability-aware exogenous variables across prediction horizons. The x-axis indicates the evaluated exogenous groups, and y-axis indicates the errors under the given group, dashed indicates the baseline performance, and labels annotate the $\Delta$ against the baseline. Lower is better.}
    \label{fig:ablation}
\end{figure}

We conducted an ablation study by incrementally augmenting the baseline with each covariate group. A Chronos-2 with calendar features as default future-known exogenous variables is employed as the baseline for computing the relative errors. The evaluated exogenous groups include past observations of sensor and equipment (+PastObs) which only consider past-known exogenous that defined in Section~\ref{sec:exo_avail}, future-known operational schedules (+OpSched), and the meteorological forecasts (+WeatherFcst).
% The Table of Figure 3 The average raw MAE
% \begin{table}[!htb]
%     \centering
%     \caption{The detailed raw MAE for availability-aware exogenous variable ablation in prediction horizon \{24, 48, 72, 96\}.}
%     \begin{tabular}{c | l | c c c c}
%         \toprule
%         & Variant & $H=24$ & $H=48$ & $H=72$ & $H=96$ \\
%         \hline
%         \multirow{4}{*}{\rotatebox{90}{\textbf{LBNL59}}}
%         & Base         & 0.519 & 0.688 & 0.759 & 0.793 \\
%         & + PastObs    & 0.489 & 0.673 & 0.765 & 0.806 \\
%         & + OpSched    & 0.481 & 0.660 & 0.750 & 0.792 \\
%         % & + WeatherFcst & 0.460 & 0.616 & 0.701 & 0.741 \\
%         & + WeatherFcst & 0.475 & 0.640 & 0.725 & 0.768 \\
%         \hline
%         \multirow{4}{*}{\rotatebox{90}{\textbf{BTS-B}}}
%         & Base         & 0.316 & 0.409 & 0.430 & 0.433 \\
%         & + PastObs    & 0.331 & 0.450 & 0.465 & 0.465 \\
%         & + OpSched    & 0.295 & 0.401 & 0.420 & 0.423 \\
%         % & + WeatherFcst & 0.246 & 0.313 & 0.312 & 0.306 \\
%         & + WeatherFcst & 0.295 & 0.388 & 0.408 & 0.413 \\
%         \hline
%         \multirow{4}{*}{\rotatebox{90}{\textbf{BTS-C}}}
%         & Base         & 0.326 & 0.427 & 0.470 & 0.495 \\
%         & + PastObs    & 0.321 & 0.431 & 0.480 & 0.506 \\
%         & + OpSched    & 0.320 & 0.430 & 0.480 & 0.507 \\
%         % & + WeatherFcst & 0.305 & 0.404 & 0.449 & 0.475 \\
%         & + WeatherFcst & 0.319 & 0.427 & 0.476 & 0.502 \\
%         \bottomrule
%     \end{tabular}
%     \label{tab:ablation_known_covs}
% \end{table}
As the visualized results in Figure~\ref{fig:ablation}, the meteorological forecasts (+WeatherFcst) yield the largest and most consistent performance improvements. Compared to the baseline, the maximum nMAE decreased by 8.5\% on LBNL59 and by 6.6\% on BTS-B, demonstrating the critical impact of future-known meteorological factors in building sensing. Operational schedules (+OpSched) ranked second, offering improvements of up to 7.3\%, which reflect the tight coupling between HVAC commands/setpoints and observed dynamics. In contrast, past observations of sensor and equipment (+PastObs) provided minimal benefit and even had a detrimental effect in BTS-B. Our analysis suggests that because these signals are masked during the prediction horizon, enriching historical context without providing future information, zero-shot models cannot always reliably leverage them. Furthermore, the BTS-C dataset under a long horizon does not seem to respond well to our ablation modules, which will be discussed in Section~\ref{sec:discussion}. 
This experiment also suggests that the information gains are mainly from future-known exogenous factors, especially meteorological forecasts, highlighting the value of modeling formulations that account for signal availability. \\

\subsubsection{Sensitivity to Meteorological Forecast Quality}
\begin{figure}
    \centering
    \includegraphics[width=\linewidth]{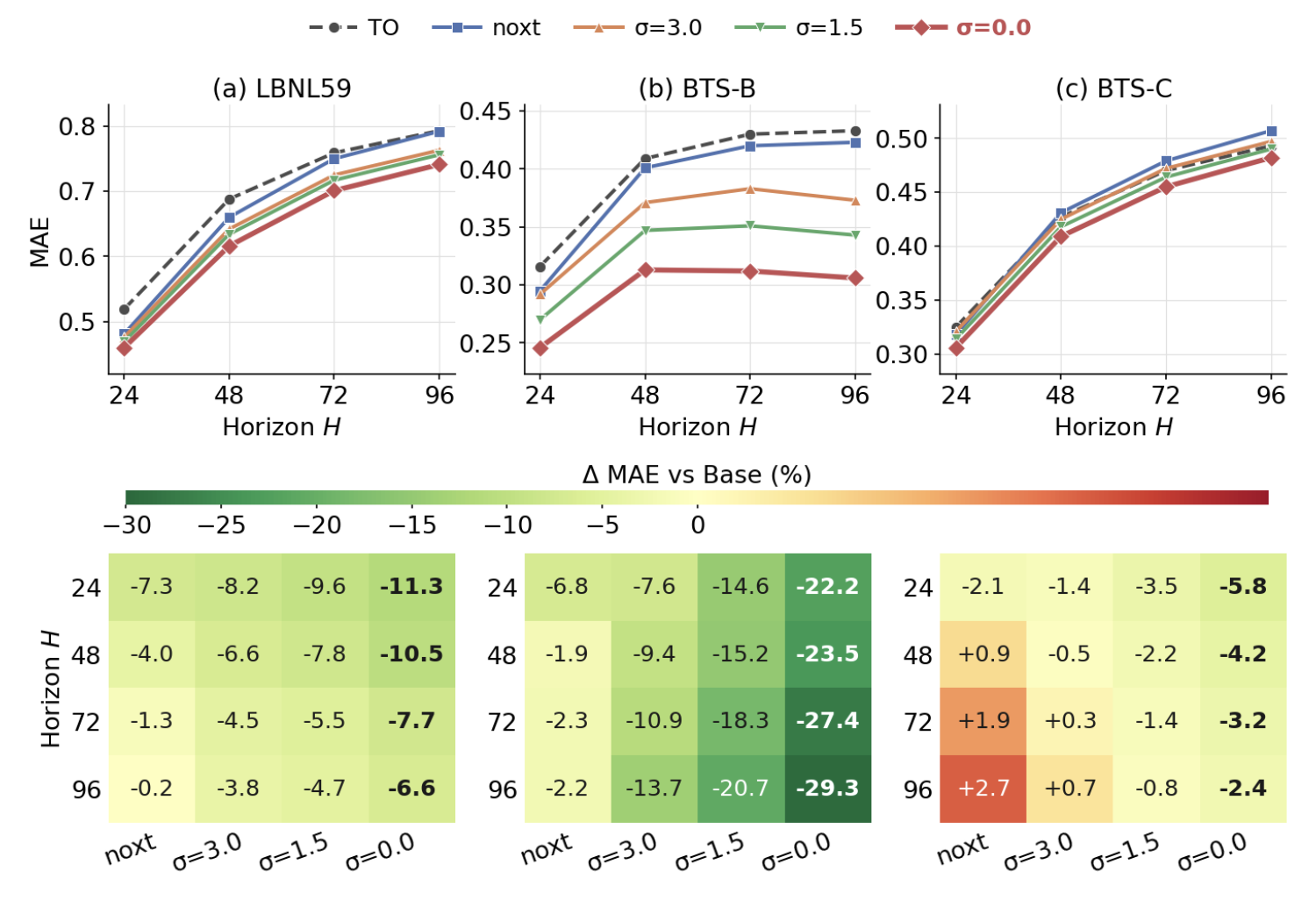}
    \caption{Sensitivity analysis of model performance to meteorological forecast quality.}
    \label{fig:external_futkn_impact}
\end{figure}

This experiment evaluates the performance sensitivity towards the quality of meteorological forecasts by injecting controllable Gaussian noise $\sigma$ into the oracle meteorological readings as future-known exogenous variables. The smaller the $\sigma$, the higher the meteorological forecast quality, where $\sigma=0$ is the oracle. 
The results in Figure~\ref{fig:external_futkn_impact} show a consistent monotonic trend: as meteorological forecast quality improves, forecasting error decreases. This confirms that weather is a useful future-known exogenous signal. On LBNL59, the oracle meteorological setting reduces nMAE by 11.3\%, 10.5\%, 7.7\%, and 6.6\% from 24 to 96 steps, respectively. Even with noisy meteorological inputs, the model remains better than the non-weather variant, indicating that the framework does not require perfect weather forecasts to be beneficial. On BTS-B, with the oracle meteorological inputs, the improvement increases from 22.2\% at 24 steps to 29.3\% at 96 steps. This suggests that meteorological conditions provide strong long-horizon explanatory power for this building. Under the noisy setting $\sigma=3.0$, BTS-B still obtains consistent reductions across all horizons, showing that the method is robust to realistic forecast uncertainty. BTS-C shows a weaker but still informative pattern. Without meteorological forecasts (noxt), the model slightly degrades at longer horizons. Adding noisy weather already recovers part of this loss, and oracle weather further improves the results. This suggests that BTS-C is less weather-dominated and potentially more affected by unobserved occupancy, equipment usage, or closed-loop control dynamics. Nevertheless, the monotonic improvement from noisy to oracle weather confirms that meteorological quality remains relevant.

% Overall, this ablation demonstrates that the proposed availability-aware formulation is deployment-sensitive: future-known meteorological exogenous variables improve forecasting accuracy, and better forecast quality leads to stronger gains. At the same time, the positive results under noisy weather indicate that TopoBrick is not an oracle-only method; it can still benefit from imperfect meteorological forecasts available in practical deployment.

% \begin{figure*}[t]
%     \centering
%     \begin{subfigure}[t]{0.82\linewidth}
%         \centering
%         \includegraphics[width=\linewidth]{figure/lbnl59_ext_fut_gains.png}
%         \caption{LBNL59}
%         \label{fig:extfut_gains_lbnl59}
%     \end{subfigure}

%     \vspace{0.5em}

%     \begin{subfigure}[t]{0.82\linewidth}
%         \centering
%         \includegraphics[width=\linewidth]{figure/btsb_ext_fut_gains.png}
%         \caption{BTS-B}
%         \label{fig:extfut_gains_btsb}
%     \end{subfigure}
%     \caption{Performance analysis by external forecasting quality. The 4 variates indicates the model performance under }
%     \label{fig:delta_analysis_by_extfut_quality}
% \end{figure*}

\section{Discussion}
\label{sec:discussion}
In heterogeneous buildings, the available exogenous points for a given target are numerous, unevenly distributed, and partially relevant to the given target. An arbitrary selection of exogenous points can degrade performance, while fixed radius graph neighborhoods are semantically unaware. The main bottleneck is the lack of a target-specific mechanism to decide the relevance.  
TopoBrick uses the building KG to impose this routing prior: the graph does not directly predict the future, but constrains what the forecaster should condition on.

The results characterize the building KG primarily as a routing prior: it constrains what the forecaster conditions on rather than predicting future values directly. However, its benefits depend on two conditions: whether the target's dominant drivers are observable in the graph and whether the selected signals are available at prediction time. The ontology-level results in Section~\ref{sec:exp_ontology} reveal this pattern. Targets governed by instrumented air-side, water-side, and thermodynamic interactions generally improve, whereas electrical power, demand, airflow, and static pressure are often limited.

The first condition defines the coverage boundary of topology-based routing. The latter quantities are strongly influenced by occupancy, discretionary equipment use, and supervisory control logic, none of which is represented as a Point in the evaluated buildings. No routing strategy can recover a driver that is neither measured nor represented in the KG. In such cases, the sampler may return structurally nearby but weakly predictive variables, diluting the information already contained in the target's history. BTS-C illustrates this limitation most clearly. Addressing it requires richer sensing and metadata coverage, rather than simply expanding the graph neighborhood or using a larger forecaster.

The second condition is deployment-time availability. Past-known sensor and equipment states enrich the historical context but provide no information within the prediction horizon, limiting their value, particularly for longer forecasts. Future-known variables, such as operational schedules and meteorological forecasts, provide more consistent gains because they remain available throughout the horizon. The weather-quality analysis further shows that improved forecasts yield larger gains, while imperfect forecasts can still be useful. Relevant context must therefore be selected jointly according to physical relevance and prediction-time availability.

The formulation may extend beyond forecasting. Virtual sensing, sensor recovery, fault localization, imputation, and control-oriented reasoning all require identifying Points related to a target through building structure, and evaluating topology-routed context there is left to future work.

\section{Conclusion}
\label{sec:conclusion}
This work proposes TopoBrick, a training-free framework for zero-shot building IoT forecasting. It samples physically coupled exogenous variables for each target by agentic reasoning over the building knowledge graph and organizes them by deployment-time availability, letting a frozen foundation model condition on meaningful context without building-specific training. On three real-world buildings TopoBrick outperforms zero-shot foundation models in most cases and remains competitive with fully trained building-specific forecasters, with benefits that depend on how closely the routed variables match the building's physical coupling. Building knowledge graphs can therefore serve as effective routing layers for forecasting across heterogeneous cyber-physical environments.

\section*{Acknowledgment}
\label{sec:ack}
This research is funded by the NSW Government through CSIRO’s NSW Digital Infrastructure Energy Flexibility (DIEF) project as part of the Net Zero Plan Stage 1: 2020-2030, and by the Reliable Affordable Clean Energy for 2030 (RACE for 2030) Cooperative Research Centre.

% Flush pending floats so tables/figures don't drift into the appendix
\FloatBarrier
\clearpage

\appendix

\bibliographystyle{ACM-Reference-Format}
\bibliography{0references}

@article{luo2022three,
  title={A three-year dataset supporting research on building energy management and occupancy analytics},
  author={Luo, Na and Wang, Zhe and Blum, David and Weyandt, Christopher and Bourassa, Norman and Piette, Mary Ann and Hong, Tianzhen},
  journal={Scientific data},
  volume={9},
  number={1},
  pages={156},
  year={2022},
  publisher={Nature Publishing Group UK London}
}

@inproceedings{zhang2023mlpst,
  title={Mlpst: Mlp is all you need for spatio-temporal prediction},
  author={Zhang, Zijian and Huang, Ze and Hu, Zhiwei and Zhao, Xiangyu and Wang, Wanyu and Liu, Zitao and Zhang, Junbo and Qin, S Joe and Zhao, Hongwei},
  booktitle={Proceedings of the 32nd ACM International Conference on Information and Knowledge Management},
  pages={3381--3390},
  year={2023}
}

@inproceedings{
xia2025timeemb,
title={TimeEmb: A Lightweight Static-Dynamic Disentanglement Framework for Time Series Forecasting},
author={Mingyuan Xia and Chunxu Zhang and Zijian Zhang and Hao Miao and Qidong Liu and Yuanshao Zhu and Bo Yang},
booktitle={NeurIPS},
year={2025},
url={https://openreview.net/forum?id=sLfMvrkn6T}
}

@inproceedings{yin2024enhancing,
  title={Enhancing spatio-temporal quantile forecasting with curriculum learning: Lessons learned},
  author={Yin, Du and Deng, Jinliang and Ao, Shuang and Li, Zechen and Xue, Hao and Prabowo, Arian and Jiang, Renhe and Song, Xuan and Salim, Flora},
  booktitle={Proceedings of the 32nd ACM International Conference on Advances in Geographic Information Systems},
  pages={42--53},
  year={2024}
}

@inproceedings{yin2025xxltraffic,
  title={XXLTraffic: Expanding and Extremely Long Traffic forecasting beyond test adaptation},
  author={Yin, Du and Xue, Hao and Prabowo, Arian and Ao, Shuang and Salim, Flora},
  booktitle={Proceedings of the 33rd ACM International Conference on Advances in Geographic Information Systems},
  pages={511--521},
  year={2025}
}

@article{prabowo2024building,
  title={Building Timeseries Dataset: Empowering Large-Scale Building Analytics},
  author={Prabowo, Arian and Lin, Xiachong and Razzak, Imran and Xue, Hao and Yap, Emily W and Amos, Matthew and Salim, Flora D},
  journal={Advances in Neural Information Processing Systems},
  volume={37},
  pages={133180--133206},
  year={2024}
}

@article{nie2022time,
  title={A time series is worth 64 words: Long-term forecasting with transformers},
  author={Nie, Yuqi and Nguyen, Nam H and Sinthong, Phanwadee and Kalagnanam, Jayant},
  journal={arXiv preprint arXiv:2211.14730},
  year={2022}
}

@inproceedings{zeng2023transformers,
  title={Are transformers effective for time series forecasting?},
  author={Zeng, Ailing and Chen, Muxi and Zhang, Lei and Xu, Qiang},
  booktitle={Proceedings of the AAAI conference on artificial intelligence},
  volume={37},
  number={9},
  pages={11121--11128},
  year={2023}
}

@article{liu2024autotimes,
  title={Autotimes: Autoregressive time series forecasters via large language models},
  author={Liu, Yong and Qin, Guo and Huang, Xiangdong and Wang, Jianmin and Long, Mingsheng},
  journal={Advances in Neural Information Processing Systems},
  volume={37},
  pages={122154--122184},
  year={2024}
}

@article{wang2024timexer,
  title={Timexer: Empowering transformers for time series forecasting with exogenous variables},
  author={Wang, Yuxuan and Wu, Haixu and Dong, Jiaxiang and Qin, Guo and Zhang, Haoran and Liu, Yong and Qiu, Yunzhong and Wang, Jianmin and Long, Mingsheng},
  journal={Advances in Neural Information Processing Systems},
  volume={37},
  pages={469--498},
  year={2024}
}

@inproceedings{zhou2021informer,
  title={Informer: Beyond efficient transformer for long sequence time-series forecasting},
  author={Zhou, Haoyi and Zhang, Shanghang and Peng, Jieqi and Zhang, Shuai and Li, Jianxin and Xiong, Hui and Zhang, Wancai},
  booktitle={Proceedings of the AAAI conference on artificial intelligence},
  volume={35},
  number={12},
  pages={11106--11115},
  year={2021}
}

@inproceedings{liu2024itransformer,
  title={itransformer: Inverted transformers are effective for time series forecasting},
  author={Liu, Yong and Hu, Tengge and Zhang, Haoran and Wu, Haixu and Wang, Shiyu and Ma, Lintao and Long, Mingsheng},
  booktitle={International conference on learning representations},
  volume={2024},
  pages={11116--11140},
  year={2024}
}

@article{zhou2023one,
  title={One fits all: Power general time series analysis by pretrained lm},
  author={Zhou, Tian and Niu, Peisong and Sun, Liang and Jin, Rong and others},
  journal={Advances in neural information processing systems},
  volume={36},
  pages={43322--43355},
  year={2023}
}

@article{ansari2024chronos,
  title={Chronos: Learning the language of time series},
  author={Ansari, Abdul Fatir and Stella, Lorenzo and Turkmen, Caner and Zhang, Xiyuan and Mercado, Pedro and Shen, Huibin and Shchur, Oleksandr and Rangapuram, Syama Sundar and Arango, Sebastian Pineda and Kapoor, Shubham and others},
  journal={arXiv preprint arXiv:2403.07815},
  year={2024}
}

@inproceedings{jin2023time,
  title={{Time-LLM}: Time series forecasting by reprogramming large language models},
  author={Jin, Ming and Wang, Shiyu and Ma, Lintao and Chu, Zhixuan and Zhang, James Y and Shi, Xiaoming and Chen, Pin-Yu and Liang, Yuxuan and Li, Yuan-Fang and Pan, Shirui and Wen, Qingsong},
  booktitle={International Conference on Learning Representations (ICLR)},
  year={2024}
}

@article{ansari2025chronos,
  title={Chronos-2: From Univariate to Universal Forecasting},
  author={Ansari, Abdul Fatir and Shchur, Oleksandr and K{\"u}ken, Jaris and Auer, Andreas and Han, Boran and Mercado, Pedro and Rangapuram, Syama Sundar and Shen, Huibin and Stella, Lorenzo and Zhang, Xiyuan and others},
  journal={arXiv preprint arXiv:2510.15821},
  year={2025}
}

@inproceedings{wang2024timemixer,
  title={Timemixer: Decomposable multiscale mixing for time series forecasting},
  author={Wang, Shiyu and Wu, Haixu and Shi, Xiaoming and Hu, Tengge and Luo, Huakun and Ma, Lintao and Zhang, James and Zhou, Jun},
  booktitle={International conference on learning representations},
  volume={2024},
  pages={38626--38652},
  year={2024}
}

@inproceedings{mulayim2025buildingqa,
  title={BuildingQA: A Benchmark for Natural Language Question Answering over Building Knowledge Graphs},
  author={Mulayim, Ozan Baris and Anwar, Avia and Saka, Umut Mete and Paul, Lazlo and Prakash, Anand Krishnan and Fierro, Gabe and Pritoni, Marco and Berg{\'e}s, Mario},
  booktitle={Proceedings of the 12th ACM International Conference on Systems for Energy-Efficient Buildings, Cities, and Transportation},
  pages={65--75},
  year={2025}
}

@inproceedings{lin2024bitsa,
  title={Bitsa: Leveraging time series foundation model for building energy analytics},
  author={Lin, Xiachong and Prabowo, Arian and Razzak, Imran and Xue, Hao and Amos, Matthew and Behrens, Sam and Salim, Flora D},
  booktitle={2024 IEEE International Conference on Data Mining Workshops (ICDMW)},
  pages={891--894},
  year={2024},
  organization={IEEE}
}

@inproceedings{prabowo2025brick,
  title={Brick-by-Brick: Cyber-Physical Building Data Classification Challenge},
  author={Prabowo, Arian and Lin, Xiachong and Razzak, Imran and Xue, Hao and Amos, Matthew and White, Stephen D and Salim, Flora D},
  booktitle={Companion Proceedings of the ACM on Web Conference 2025},
  pages={3021--3025},
  year={2025}
}

@inproceedings{mulayim2025towards,
  title={Towards Zero-shot Question Answering in CPS-IoT: Large Language Models and Knowledge Graphs},
  author={Mulayim, Ozan Baris and Fierro, Gabe and Berg{\'e}s, Mario and Pritoni, Marco},
  booktitle={Proceedings of the 2nd International Workshop on Foundation Models for Cyber-Physical Systems \& Internet of Things},
  pages={7--12},
  year={2025}
}

@inproceedings{zaman2025bloom,
  title={Bloom-LLM: Privacy-Preserving Large Language Model for Load Forecasting},
  author={Zaman, Zakia and Gauravaram, Praveen and Jha, Sanjay and Hu, Wen},
  booktitle={Proceedings of the 12th ACM International Conference on Systems for Energy-Efficient Buildings, Cities, and Transportation},
  pages={128--138},
  year={2025}
}

@inproceedings{balaji2016brick,
  title={Brick: Towards a unified metadata schema for buildings},
  author={Balaji, Bharathan and Bhattacharya, Arka and Fierro, Gabriel and Gao, Jingkun and Gluck, Joshua and Hong, Dezhi and Johansen, Aslak and Koh, Jason and Ploennigs, Joern and Agarwal, Yuvraj and others},
  booktitle={Proceedings of the 3rd ACM International Conference on Systems for Energy-Efficient Built Environments},
  pages={41--50},
  year={2016}
}

@inproceedings{prabowo2023navigating,
  title={Navigating Out-of-Distribution Electricity Load Forecasting during COVID-19: Benchmarking energy load forecasting models without and with continual learning},
  author={Prabowo, Arian and Chen, Kaixuan and Xue, Hao and Sethuvenkatraman, Subbu and Salim, Flora D},
  booktitle={Proceedings of the 10th ACM international conference on systems for energy-efficient buildings, cities, and transportation},
  pages={41--50},
  year={2023}
}

@article{xue2023promptcast,
  title={Promptcast: A new prompt-based learning paradigm for time series forecasting},
  author={Xue, Hao and Salim, Flora D},
  journal={IEEE Transactions on Knowledge and Data Engineering},
  volume={36},
  number={11},
  pages={6851--6864},
  year={2023},
  publisher={IEEE}
}

@inproceedings{jia2024gpt4mts,
  title={Gpt4mts: Prompt-based large language model for multimodal time-series forecasting},
  author={Jia, Furong and Wang, Kevin and Zheng, Yixiang and Cao, Defu and Liu, Yan},
  booktitle={Proceedings of the AAAI Conference on Artificial Intelligence},
  volume={38},
  number={21},
  pages={23343--23351},
  year={2024}
}

@inproceedings{xu2024fits,
  title={FITS: Modeling time series with $10 k $ parameters},
  author={Xu, Zhijian and Zeng, Ailing and Xu, Qiang},
  booktitle={International Conference on Learning Representations},
  volume={2024},
  pages={26295--26318},
  year={2024}
}

@article{liu2025moirai,
  title={Moirai 2.0: When less is more for time series forecasting},
  author={Liu, Chenghao and Aksu, Taha and Liu, Juncheng and Liu, Xu and Yan, Hanshu and Pham, Quang and Savarese, Silvio and Sahoo, Doyen and Xiong, Caiming and Li, Junnan},
  journal={arXiv preprint arXiv:2511.11698},
  year={2025}
}

@inproceedings{woo2024unified,
  title={Unified training of universal time series forecasting transformers},
  author={Woo, Gerald and Liu, Chenghao and Kumar, Akshat and Xiong, Caiming and Savarese, Silvio and Sahoo, Doyen},
  booktitle={Forty-first International Conference on Machine Learning},
  year={2024}
}

@article{das2023decoder,
  title={A decoder-only foundation model for time-series forecasting},
  author={Das, Abhimanyu and Kong, Weihao and Sen, Rajat and Zhou, Yichen},
  journal={arXiv preprint arXiv:2310.10688},
  year={2023}
}

@inproceedings{huang2026reliable,
  title={Reliable Indoor Air Quality Prediction with Uncertainty-Aware Spatiotemporal GNN},
  author={Huang, Cong and Cheng, Jack CP and Hou, Fangli and Wu, Zhaoji and Kwok, Helen HL},
  booktitle={Proceedings of the 13th ACM International Conference on Systems for Energy-Efficient Buildings, Cities, and Transportation},
  pages={324--333},
  year={2026}
}

@article{xu2022probabilistic,
  title={Probabilistic electrical load forecasting for buildings using Bayesian deep neural networks},
  author={Xu, Lei and Hu, Maomao and Fan, Cheng},
  journal={Journal of Building Engineering},
  volume={46},
  pages={103853},
  year={2022},
  publisher={Elsevier}
}

@inproceedings{hu2025lightllm,
  title={Lightllm: A versatile large language model for predictive light sensing},
  author={Hu, Jiawei and Jia, Hong and Hassan, Mahbub and Yao, Lina and Kusy, Brano and Hu, Wen},
  booktitle={Proceedings of the 23rd ACM Conference on Embedded Networked Sensor Systems},
  pages={158--171},
  year={2025}
}

@article{lee2026jarvis,
  title = {JARVIS for HVAC: LLM-based Question-Answer Framework for Sensor-driven HVAC System Interaction},
  author = {Lee, Sungmin and Lee, Joonhee and Kang, Minju and Lee, Seungyong and Kim, Dongju and Hong, Jingi and Shin, Jun and Zhang, Pei and Ko, JeongGil},
  journal = {Proceedings of the ACM on Interactive, Mobile, Wearable and Ubiquitous Technologies},
  volume = {10},
  number = {2},
  pages = {1--36},
  year = {2026},
  publisher = {ACM New York, NY, USA},
}

@inproceedings{ko2026brickqa,
  title = {BrickQA: Bridging the Semantic Gap in Building Operations with Dynamic Graph Exploration},
  author = {Ko, Yun-Dam and Jung, Hyeyun Eunice and Pritoni, Marco and Jain, Rishee},
  booktitle = {Proceedings of the 13th ACM International Conference on Systems for Energy-Efficient Buildings, Cities, and Transportation},
  pages = {207--217},
  year = {2026},
}

@inproceedings{ma2026mixlinear,
  title={MixLinear: Extreme Low Resource Multivariate Time Series Forecasting with $0.1 K $ Parameters},
  author={Ma, Aitian and Luo, Dongsheng and Sha, Mo},
  booktitle={International Conference on Learning Representations},
  volume={2026},
  pages={77764--77789},
  year={2026}
}
\newpage

\section{Preprocessing and Forecast Target Construction}
\label{appendix:preprocess}
We design a four-stage pre-processing procedure to separate the failure modes that would be conflated by a single filter, avoiding the discard of a sensor with sparse local corruption.

\paragraph{L1: Observation-level range filtering.} Each observation $(i,t)$ carries a quality-control mask $m^{\mathrm{qc}}_{i,t}\in\{0,1\}$. A reading is masked, without removing its sensor, when it is \texttt{NaN} or infinite, matches a known sentinel code, exceeds an overflow threshold of $10^9$, falls outside the physically plausible range for its Brick class, or is flagged as an abnormal drop to zero, a frozen run, or a robust MAD outlier. Temperature-like sensors are canonicalized to degrees Celsius before their bounds are applied.

\paragraph{L2: Sensor-level usability auditing.} Each sensor then receives a usability flag $u_i$, and the exogenous pool is $\mathcal{U}_{\mathrm{exo}} = \{\, i \mid u_i = 1 \,\}$. A sensor leaves the pool when its cleaned signal is still statistically unreliable, which covers post-QC coverage below $0.4$, near-flat dynamics, almost-all-zero measurements, and monotonic cumulative-meter behaviour. Where L1 masks readings, L2 removes sensors.

\paragraph{L3: Semantic and empirical target filtering.} A sensor is kept as a forecast target when $M^{\mathrm{target}}_i = M^{\mathrm{meas}}_i \wedge M^{\mathrm{eligible}}_i \wedge u_i$, where $M^{\mathrm{meas}}_i$ marks it as a measurement Point and $M^{\mathrm{eligible}}_i$ excludes classes unsuited to continuous prediction: control and configuration variables, raw electrical identifiers, actuators, counters, setpoints, status and command points, and unmapped entities.

\paragraph{L4: Window-level validation.} Sensors whose training-split variation falls below $\mathrm{std}_{\mathrm{train}} = 0.1$ are dropped, and a window ending at $t$ is kept only when $\frac{1}{H}\sum_{\tau=t+1}^{t+H} m^{\mathrm{qc}}_{i,\tau} \ge 0.95$. This removes outage-dominated windows while keeping sensors that are usable in other periods.

% \section{Baseline Implementation Details}
% \label{appendix:impl_details}
% All supervised models are trained with Adam for up to 30 epochs under a masked MSE loss read on the same
% per-sensor normalized scale as the metric, with gradient clipping at $\lVert\cdot\rVert_{2}\!\le\!1$, early stopping with patience 5 is employed.
% In the multivariate setting, each sample is indexed by an anchor timestamp and contains a lookback window for every sensor. If a sensor does not yield a valid window at that timestamp, its entire channel is zero-filled. Although absent channels are excluded from the loss and evaluation metrics, they would still enter the model's variate-attention layer. Because iTransformer encodes each channel's full lookback window as a single token, zero-filled channels produce identical placeholder tokens. When many channels are absent, these duplicate tokens can collectively absorb a substantial share of the softmax mass, diluting attention to observed channels.
% We therefore mask absent channels on the key side of variate attention, preventing them from contributing to any channel's representation. Their query-side outputs remain harmless because they are excluded from both the loss and evaluation metrics. All other implementation details remain unchanged.
\section{Baseline Implementation Details}
\label{appendix:impl_details}
All supervised baselines use Adam for up to 30 epochs with masked MSE on the per-sensor normalized scale, gradient clipping at $\lVert\cdot\rVert_2 \le 1$, and early stopping with patience 5. Invalid multivariate windows are zero-filled and excluded from loss and evaluation. For multivariate settings, absent channels are additionally masked as attention keys to prevent duplicate zero tokens from absorbing softmax mass. All other settings remain unchanged.

\section{Pseudocode}
\label{appendix:sampler_algorithm}

Algorithm~\ref{alg:agentic_subgraph_sampling} shows the pipeline pseudocode for Section~\ref{sec:agentic_sampler} and Section~\ref{sec:physics_verifier}. 
Each target uses two agent calls: sampling and in-place verification. Verifier removals precede additions, after which the selection is capped at $n_{\max}=60$. The cap applies to $25.6\%$ of BTS-C targets and at most $2\%$ in the other buildings.

\begin{algorithm}[t]
\caption{Agentic topology subgraph sampling.}
\label{alg:agentic_subgraph_sampling}
\small
\begin{algorithmic}[1]
\Require target Point $p$; skeleton $G_s=(V_s,E_s)$; size cap $n_{\max}$
\Ensure exogenous-variable set $\mathcal{E}(p)$ and its subgraph $G(p)$

\Statex \textbf{Phase 1: deterministic context construction}
\State $A \gets \{u\in V_s \mid (u,\texttt{hasPoint},p)\in E\}$ \Comment{target anchors}
\If{$A=\emptyset$} \Comment{$p$ has no skeleton anchor}
  \State $U \gets$ recovered equipment cluster of $p$
\Else
  \State $U \gets \bigcup_{a\in A}\textsc{Upstream}(a,G_s)$ \Comment{to the top of $G_s$}
\EndIf
\State $(\mathcal{C}(p),\alpha)\gets\textsc{Context}(p,U,G_s)$

\Statex \textbf{Phase 2: topology sampler --- one agent call}
\State $\mathcal{A}(p)\gets\textsc{Sample}(\mathcal{C}(p))$ \Comment{actions $(a_i,\omega_i,q_i)$}
\State $\mathcal{L}(p)\gets\textsc{Materialize}(\mathcal{A}(p),\alpha)$ \Comment{deduplicated, $p$ excluded}

\Statex \textbf{Phase 3: physics-consistency verifier --- one agent call}
\If{the verifier is enabled}
  \State $(\mathcal{D},\mathcal{A}^{+})\gets\textsc{Verify}\big(\mathcal{C}(p),\ \mathrm{classes}(\mathcal{L}(p))\big)$
  \State $\mathcal{L}(p)\gets\big[\,\ell\in\mathcal{L}(p) \mid \mathrm{class}(\ell)\notin\mathcal{D}\,\big]$ \Comment{removals first}
  \State append $\textsc{Materialize}(\mathcal{A}^{+},\alpha)\setminus\mathcal{L}(p)$ to $\mathcal{L}(p)$
\EndIf

\Statex \textbf{Phase 4: deterministic assembly}
\State $\mathcal{E}(p)\gets\mathcal{L}(p)[1{:}n_{\max}]$
\State $G(p)\gets$ subgraph on $U\cup\{p\}\cup\mathcal{E}(p)$, one \texttt{hasPoint}
       edge per selected Point
\State \Return $\mathcal{E}(p)$, $G(p)$
\end{algorithmic}
\end{algorithm}

\section{Prompts and Output Examples for the Agents}
\label{appendix:prompt_sampler}
\paragraph{Prompts for Topology Sampler}
\label{app:topology_sampler}
The box below gives the system prompt used by the topology sampler.
\begin{tcolorbox}[
  enhanced,
  breakable,
  width=\columnwidth,
  colback=black!3,
  colframe=black!50,
  boxrule=0.4pt,
  arc=1pt,
  % tighten vertical whitespace around and inside the box
  before skip=3pt,
  after skip=3pt,
  boxsep=2pt,
  left=3pt,right=3pt,top=3pt,bottom=3pt,
  title={\textbf{Sampler System Prompt}},
  fonttitle=\bfseries
]
\begin{Verbatim}[fontsize=\scriptsize,breaklines=true,breakanywhere=true]
You are a building-systems engineer choosing covariate time series to help a forecasting model predict a target sensor. The target can be any kind of point: temperature, electrical power, energy, pressure, flow, humidity, status, or another Brick point class. Reason about this target's own physics.

Reason explicitly before answering, then output exactly one fenced JSON block.

Selection rubric:

1. Target physics. Identify what the target measures and what physically drives that quantity. Drivers are domain-specific. For example:
   - temperature: upstream supply, return, or mixed-air temperatures, setpoints, and weather;
   - electrical power/current/energy: voltage, current, metered load, and building demand;
   - flow/pressure: upstream fan, pump, valve, and the loop it sits on;
   - humidity/CO2: ventilation, occupancy, and outdoor humidity.

2. Topology localization. Read the target node, its serving or containing equipment or group, and its upstream chain.

3. Co-located state. Include relevant points directly attached to the target's own node, such as setpoints, controls, statuses, or co-located sensors.

4. Same-class peer cohort. Choose the tightest physically meaningful anchor, such as the immediate serving equipment, meter, circuit, or zone group. Prefer a small same-serving-equipment cohort over a building-wide same-class dump. Use building-wide expansion only when the local cohort is too small.

5. Domain drivers. Pull the drivers identified in Step 1 from the own node, upstream chain, or global driver hosts. Do not pull variables that are not physically coupled to the target's domain.

Self-check:
- Include the target's own controls or co-located points when relevant.
- Include a tight same-class peer cohort rather than a building-wide flood.
- Include the domain drivers identified above when they are available.
- Reject off-domain variables with no coupling pathway, static configuration points, and duplicates.

Context format:
- YOU ARE HERE contains the target node and its ancestry to the root. feeds denotes upstream flow or supply, including the circuit that powers it. hasPart and isLocationOf denote spatial containment.
- NEIGHBORHOOD contains anchors [A#], their directly attached point classes, and their direct child equipment or zones.
- GLOBAL DRIVERS contains building-wide hosts such as weather stations, utility meters, central plant, and loops.

Each pull action has the form:
{
  "anchor": "A#",
  "expand": "self" | "subtree" | "building" | "cluster",
  "point_class": "<Brick point class>",
  "reason": "short rationale"
}

Expansion semantics:
- self: points of the requested class directly attached to the anchor.
- subtree: points of the requested class under the anchor subtree.
- building: points of the requested class across the entire building. Use only for scarce global signals, not common classes that appear on many unrelated equipment.
- cluster: recovered cluster mates for KG-orphan targets.

Return exactly:
{
  "pulls": [
    {
      "anchor": "A#",
      "expand": "self" | "subtree" | "building" | "cluster",
      "point_class": "<Brick point class>",
      "reason": "..."
    }
  ],
  "summary": "one-line summary"
}
\end{Verbatim}
\end{tcolorbox}

\paragraph{Prompts for Verifier}
The box below gives the system prompt used by the physics-consistency verifier.
\begin{tcolorbox}[
  enhanced,
  breakable,
  width=\columnwidth,
  colback=black!3,
  colframe=black!50,
  boxrule=0.4pt,
  arc=1pt,
  % tighten vertical whitespace around and inside the box
  before skip=3pt,
  after skip=3pt,
  boxsep=2pt,
  left=3pt,right=3pt,top=3pt,bottom=3pt,
  title={\textbf{Verifier System Prompt}},
  fonttitle=\bfseries
]
\begin{Verbatim}[fontsize=\scriptsize,breaklines=true,breakanywhere=true]
You are an independent senior building-systems engineer auditing another engineer's covariate selection for a forecasting target. You are shown the same building neighborhood and the selected covariate set summarized by Brick class and count.

Audit only for this target's physics:

1. Peer cohort. Check whether same-class spatial peers from the target's serving equipment or group are included. If such peers clearly exist nearby but are missing or under-represented, flag this as a gap.

2. Domain drivers. Check whether available physical drivers of the target quantity are present: weather for thermal variables; voltage/current for power; upstream fan, pump, valve, or flow variables for pressure/flow; and ventilation or occupancy variables for humidity/CO2.

3. Noise. Check whether the selection contains off-domain classes without a plausible coupling pathway, or an excessively broad same-class dump that should be trimmed.

Most selections are already acceptable. Flag only real gaps or excesses, and only use variables actually available in the shown neighborhood.

Return exactly one JSON object:
{
  "verdict": "ok" | "revise",
  "add": [
    {
      "anchor": "A#",
      "expand": "self" | "subtree" | "building" | "cluster",
      "point_class": "...",
      "reason": "missing peer" | "missing driver"
    }
  ],
  "drop": ["BrickClass_to_remove", "..."],
  "note": "one-line summary"
}

If the selection is already good, return:
{"verdict": "ok", "add": [], "drop": [], "note": "..."}
\end{Verbatim}
\end{tcolorbox}

\paragraph{Sampled Output}
The box below shows the sampler output for a zone air temperature target
(\texttt{zone\_016\_temp}, LBNL59), reproduced verbatim from the released
subgraph artifact. The upper block is what the agent returns and conforms to the
schema of the prompt above. The lower block lists the fields that the
deterministic materializer and the verifier attach afterwards, which are recorded
alongside it but are not part of the agent's response.
\begin{tcolorbox}[
  enhanced,
  breakable,
  width=\columnwidth,
  colback=black!3,
  colframe=black!50,
  boxrule=0.4pt,
  arc=1pt,
  % tighten vertical whitespace around and inside the box
  before skip=3pt,
  after skip=3pt,
  boxsep=2pt,
  left=3pt,right=3pt,top=3pt,bottom=3pt,
  title={\textbf{Sampler Output (JSON)}},
  fonttitle=\bfseries
]
\begin{Verbatim}[fontsize=\scriptsize,breaklines=true,breakanywhere=true]
### agent response (conforms to the schema in the prompt above)
{
  "pulls": [
    {"anchor": "A1", "expand": "self",
     "point_class": "Heating_Temperature_Setpoint",
     "reason": "Direct control of zone heating setpoint"},
    {"anchor": "A1", "expand": "self",
     "point_class": "Cooling_Temperature_Setpoint",
     "reason": "Direct control of zone cooling setpoint"},
    {"anchor": "A1", "expand": "self",
     "point_class": "Supply_Air_Flow_Sensor",
     "reason": "Supply flow directly influences zone temperature"},
    {"anchor": "A3", "expand": "subtree",
     "point_class": "Supply_Air_Temperature_Sensor",
     "reason": "Upstream supply air temperature drives zone temperature"},
    {"anchor": "A3", "expand": "subtree",
     "point_class": "Mixed_Air_Temperature_Sensor",
     "reason": "Mixed air temperature is a key upstream driver"},
    {"anchor": "A4", "expand": "subtree",
     "point_class": "Zone_Air_Temperature_Sensor",
     "reason": "Peer cohort of zone temps within the same HVAC zone"},
    {"anchor": "A5", "expand": "self",
     "point_class": "Occupant_Count",
     "reason": "Occupancy load influences zone temperature"},
    {"anchor": "A6", "expand": "building",
     "point_class": "Outdoor_Air_Temperature_Sensor",
     "reason": "Outdoor temperature is a global driver for zone thermal balance"}
  ],
  "summary": "Pull zone setpoints, flow, upstream air temps, peer zone temps, occupancy, and outdoor temp"
}

### attached downstream (materializer, then verifier)
n_resolved per pull : 1, 1, 1, 1, 1, 10, 1, 2
verifier: {"verdict": "ok", "added": 0, "dropped_classes": [],
            "note": "All peer, driver, and noise checks satisfied."}
subgraph: 25 nodes, 14 edges, spine_size 6
leaves: 18 (n_pre_cap 18, below the 60-leaf cap)
roots_reached: Building, RTU01
\end{Verbatim}
\end{tcolorbox}

\section{Cost of Subgraph Construction}
\label{appendix:token_cost}

Each target requires two LLM calls, one for sampling and one for verification. Subgraphs are constructed once during building onboarding and reused across all prediction horizons. Prompt and completion denote API-reported input and output token usage, respectively.

\begin{table}[h]
\centering
\small
\setlength{\tabcolsep}{4pt}
\caption{Token usage for subgraph construction}
\label{tab:token_cost}
\begin{tabularx}{\columnwidth}{@{}l YYY@{}}
\toprule
\textbf{Building} & \textbf{Prompt} & \textbf{Completion} & \textbf{Total} \\
\midrule
LBNL59 & 2{,}412  & 1{,}353 & 3{,}765 \\
BTS-B  & 4{,}718  & 1{,}092 & 5{,}810 \\
BTS-C  & 12{,}165 & 995     & 13{,}160 \\
\bottomrule
\end{tabularx}
\end{table}

Variation in token usage is driven primarily by prompt length. Larger topology contexts increase input usage, whereas completion usage remains comparatively stable. Because subgraphs are constructed once and reused, this overhead is incurred once per target rather than per forecast. Decoding is nondeterministic, so the accounting pass may not reproduce the released subgraphs exactly.

\end{document}